\documentclass[preprint,12pt]{elsarticle}

\usepackage{amssymb}
\usepackage{amsmath}
\usepackage{graphicx}%
\usepackage{multirow}%
\usepackage{amsmath,amssymb,amsfonts}%
\usepackage{enumitem}
\usepackage{amsthm}%
\usepackage{mathrsfs}%
\usepackage[title]{appendix}%
\usepackage{xcolor}%
\usepackage{textcomp}%
\usepackage{manyfoot}%
\usepackage{booktabs}%
\usepackage{algorithm}%
\usepackage{algorithmicx}%
\usepackage{algpseudocode}%
\usepackage{listings}%
\usepackage{makecell}
\usepackage{subcaption}
\usepackage{tikz}
\usepackage[T1]{fontenc}
\usepackage{colortbl}
\usepackage{hyperref}
\journal{Information Fusion}

\begin{document}

\begin{frontmatter}



\title{The Duality of Generative AI and Reinforcement Learning in Robotics: A Review}

\author[label1]{Angelo Moroncelli\corref{cor1}} %% Author name
\ead{angelo.moroncelli@supsi.ch}
\author[label1,label3]{Vishal Soni}
\author[label1]{Marco Forgione}
\author[label1]{Dario Piga}
\author[label4]{Blerina Spahiu}
\author[label1,label5]{Loris Roveda\corref{cor1}}
\ead{loris.roveda@supsi.ch}
\ead{loris.roveda@polimi.it}

\cortext[cor1]{Corresponding author(s)}
\affiliation[label1]{organization={Department of Innovative Technologies, IDSIA, University of Applied Science and Arts of Southern Switzerland},%Department and Organization
            city={Lugano},
            postcode={6900}, 
            state={Ticino},
            country={Switzerland}}

%\affiliation[label2]{organization={Faculty of Informatics, Università della Svizzera Italiana},%Department and Organization
%            city={Lugano},
%            postcode={6900}, 
%            state={Ticino},
%            country={Switzerland}}

\affiliation[label3]{organization={Department of Mechatronics and Automation Engineering, Indian Institute of Information Technology Bhagalpur},%Department and Organization
            city={Bhagalpur},
            postcode={813210}, 
            state={Bihar},
            country={India}}

\affiliation[label4]{organization={Department of Informatics, Systems and Communication, Università di Milano Bicocca},%Department and Organization
            city={Milano},
            postcode={20126}, 
            state={Lombardy},
            country={Italy}}

\affiliation[label5]{organization={Mechanical Department, Politecnico di Milano},%Department and Organization
            city={Milano},
            postcode={20156}, 
            state={Lombardy},
            country={Italy}}

%% Abstract
\begin{abstract}
Recently, generative AI and reinforcement learning (RL) have been redefining what is possible for AI agents that take information flows as input and produce intelligent behavior. As a result, we are seeing similar advancements in embodied AI and robotics for control policy generation.
Our review paper examines the integration of generative AI models with RL to advance robotics. Our primary focus is on the duality between generative AI and RL for robotics downstream tasks. Specifically, we investigate: (1) The role of prominent generative AI tools as modular priors for multi-modal input fusion in RL tasks. (2) How RL can train, fine-tune and distill generative models for policy generation, such as VLA models, similarly to RL applications in large language models. We then propose a new taxonomy based on a considerable amount of selected papers.

Lastly, we identify open challenges accounting for model scalability, adaptation and grounding, giving recommendations and insights on future research directions. We reflect on which generative AI models best fit the RL tasks and why. On the other side, we reflect on important issues inherent to RL-enhanced generative policies, such as safety concerns and failure modes, and what are the limitations of current methods. A curated collection of relevant research papers is maintained on our \href{https://github.com/clmoro/Robotics-RL-FMs-Integration}{GitHub repository}, serving as a resource for ongoing research and development in this field.
\end{abstract}

%% Keywords

\end{frontmatter}

\section{Introduction} 
\label{sec:introduction}

Two main pursuits of embodied intelligence and robotics are achieving physical grounding~\citep{cohen2024survey,kollar2017generalized}, the capability of robots to sense, comprehend, and interact efficiently with the physical environment, and human-level reasoning on abstract concepts~\citep{cohen2024survey,carta2023grounding,ahn2022can,Huang2023GroundedDG}, which remain a central goal of artificial intelligence (AI) research in general. Recently, the convergence of two powerful paradigms, generative AI and Reinforcement Learning (RL), has shown significant promise in advancing this objective~\citep{xietext2reward,liu2024rl,wang2024,ma2023eureka,ma2024dreureka,zhu2023diffusion,Huang2023DiffusionReward}. On the one hand, tools from generative AI, including Large Language Models (LLMs), multi-modal foundation models~\citep{bommasani2021opportunities} such as Vision-Language Models (VLMs), and world or video prediction models, have excelled in processing and generating diverse data types such as text, code, and images~\citep{baumli2023vision,chen2024vision,rocamonde2023vision,hassan2024gem}. Trained on vast multi-modal datasets, these models encapsulate rich, generalizable knowledge representations~\citep{chen2023open,ha2018recurrent,qi2025gpc}. Moreover, diffusion models excel in generative capabilities and robust training processes for downstream tasks such as state representation and policy generation~\citep{he2024diffusion,chi2023diffusion}. RL, on the other hand, offers a framework for agents to learn optimal behaviors through interaction with their environment~\citep{712192,mnih2013playingatarideepreinforcement}.

The potential synergy between generative AI and RL is particularly compelling in the context of robotics. Generative models can serve as robust priors for information fusion in RL agents---\textit{i.e.}, fusing information flows from different sources to produce a robot control policy---providing extensive world knowledge, grounding language understanding, and generating rich behaviors~\citep{firoozi2023foundationmodelsroboticsapplications,ma2023eureka,wang2024}. Conversely, RL can ``embody'' generative AI models used as generative policies for robotics, usually trained through pure Imitation Learning (IL)~\citep{open_x_embodiment_rt_x_2023}, allowing them to interact with and learn from dynamic physical environments and suboptimal data~\citep{sutton1988learning}. This integration could lead to robotic systems with enhanced adaptability, generalization, and overall intelligence through feature alignment with experience~\citep{tian2024maximizing}, similarly to RL from human feedback training for LLMs~\citep{ziegler2019fine}.

Foundation models, in particular, have demonstrated exceptional proficiency in processing information for downstream robotics tasks~\citep{huang2023voxposer,shridhar2022cliport,bhat2024grounding}. Additionally, specialized foundation models for robotics exist~\citep{firoozi2023foundationmodelsroboticsapplications}. Foundation models have also been adopted in the field of learning and control of dynamical systems, where robotics represents a key area of application. Transformer-based pre-trained models have been proposed \citep{forgione2023system,du2023can} for zero-shot prediction of outputs from a class of dynamical systems in response to any query input sequence, and for in-context state estimation~\citep{busetto2024context}. These advancements highlight the potential of foundation models to introduce innovative approaches and paradigms to traditional dynamical control systems, facilitating a shift towards data-driven  estimation and control synthesis for classes of dynamical systems, rather than for single, specific systems.

However, despite significant progress in foundation models, their application with RL in robotics remains underexplored. Recent developments suggest this promising union to enhance robotic learning and generalization, yet challenges persist, especially for real-world robotic applications~\citep{ma2024surveyvisionlanguageactionmodelsembodied}.

\begin{figure}[!htbp]
    \centering
    \begin{subfigure}[t]{0.49\textwidth}
        \centering
        \includegraphics[width=\textwidth]{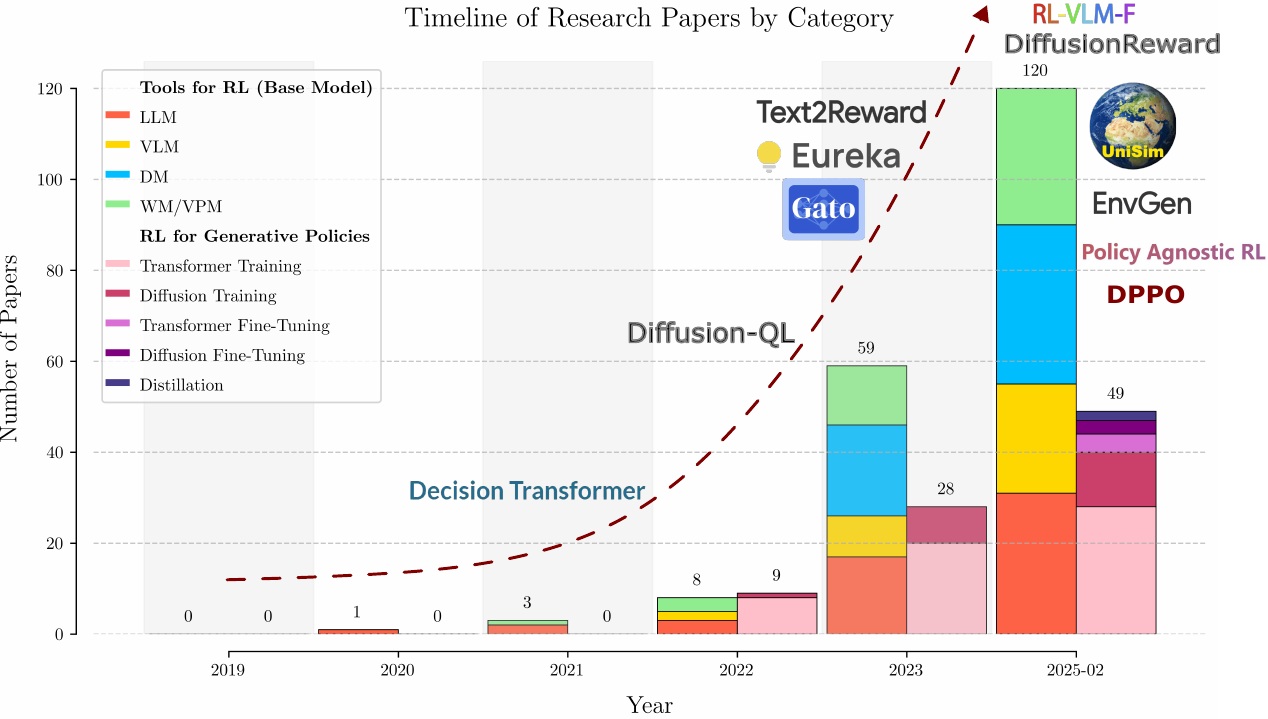}
        \captionsetup{justification=centering}
        \caption{The figure illustrates the number of papers published each year integrating both generative AI and RL in robotics, categorized by the type of model employed.}
    \label{fig:motivation_timeline}
    \end{subfigure}
    \hfill
    \begin{subfigure}[t]{0.49\textwidth}
        \centering
        \includegraphics[width=\textwidth]{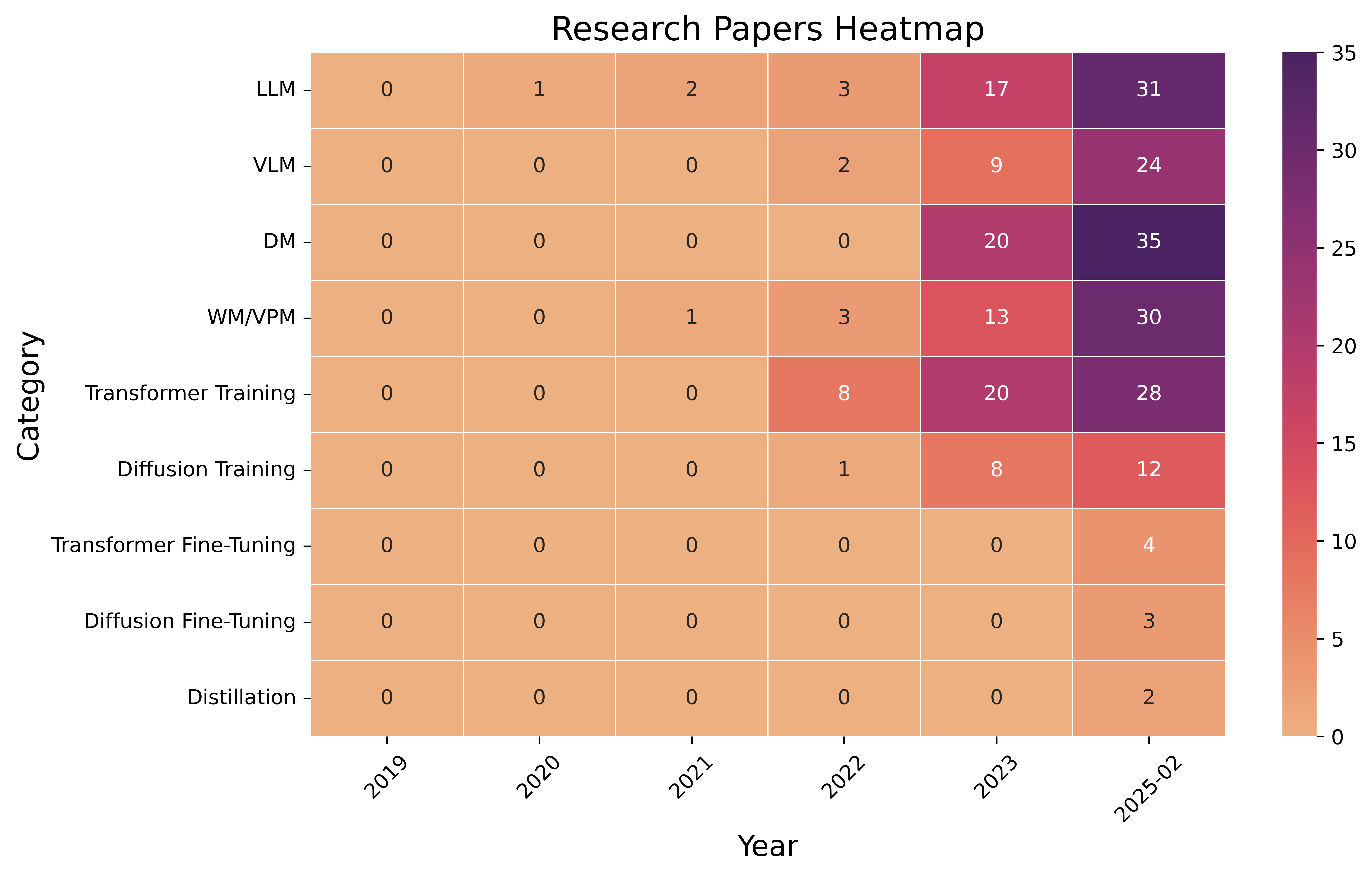}
        \captionsetup{justification=centering}
        \caption{The heatmap highlights the notable rise in the use of diffusion models as tools for RL in robotics from 2023 up to February 2025 (dark purple block). Notice the recent introduction of few seminal works in RL-based fine-tuning and distillation of generative policies.}
    \label{fig:motivation_heatmap}
    \end{subfigure}
    \captionsetup{justification=centering}
    \caption{\textbf{Trends in generative AI and RL integration for robotics.}}
    \label{fig:trends}
\end{figure}

\subsection{Scope and Contribution}

Despite rapid advancements in both generative AI and RL~\citep{dubey2024llama,achiam2023gpt,mnih2013playingatarideepreinforcement,reed2022generalist}, there is a noticeable lack of comprehensive studies systematically exploring their integration in robotics (see Table~\ref{tab:comparison}). Few research either focuses on these areas in isolation~\citep{hu2023generalpurposerobotsfoundationmodels,xiao2023robotlearningerafoundation,firoozi2023foundationmodelsroboticsapplications,zhou2024languageconditionedlearningroboticmanipulation}, or examines narrow aspects of their combination~\citep{cao2024surveylargelanguagemodelenhanced,zhu2023diffusion}. Our review addresses this gap by providing an in-depth analysis of current research trends at the intersection of generative AI and RL in robotics. Specifically, it examines Transformer- and Diffusion-based generative AI models (LLMs, VLMs, diffusion models (DMs), world models (WMs) and video prediction models (VPMs)) currently used to enhance RL, considering the diverse data modalities, roles, and applications (see Sections~\ref{sec:basemodel},~\ref{sec:modality}, and~\ref{sec:task}). Figure~\ref{fig:motivation_timeline} illustrates the growing adoption of generative AI models for RL, particularly diffusion models as tools for RL in 2023–2024, as shown in Figure~\ref{fig:motivation_heatmap}. Additionally, particular attention is given to the unique relationship between generative policies, which are a specific and relatively narrow type of generative models for robotic action generation, and their integration with RL (see Sections~\ref{sec:genpolicies},~\ref{sec:finetuning}, and~\ref{sec:distillation}), with insights on their good fitting with learning-based control techniques for grounding into downstream robotic tasks (see Section~\ref{sec:conclusions}). To the best of our knowledge, RL fine-tuning of generalist generative policies has also not been classified in previous surveys~\citep{hu2023generalpurposerobotsfoundationmodels,xiao2023robotlearningerafoundation,firoozi2023foundationmodelsroboticsapplications,zhou2024languageconditionedlearningroboticmanipulation}.

\begin{table}[h]
    \centering
    \renewcommand{\arraystretch}{1.4}
    \tiny
    \begin{tabular}{lccccccc}
        \toprule
        \textbf{Survey} & \rotatebox{0}{\textbf{RFM}} & \multicolumn{4}{c}{\textbf{Tools for RL}} & \multicolumn{2}{c}{\textbf{RL Policies}} \\
        \cmidrule(lr){3-6} \cmidrule(lr){7-8}
        & & \textbf{LLM} & \textbf{VLM} & \textbf{DM} & \textbf{WM/VPM} & \textbf{Train} & \textbf{Fine-Tune} \\
        \midrule
        
        \rowcolor{gray!10}
        \citet{hu2023generalpurposerobotsfoundationmodels} & \checkmark & \(\star\) & \(\star\) & & & \checkmark & \\
        
        \citet{xiao2023robotlearningerafoundation} & \checkmark & \(\star\) & & & & \(\star\) & \\
        
        \rowcolor{gray!10}
        \citet{firoozi2023foundationmodelsroboticsapplications} & \checkmark & \(\star\) & \(\star\) & & & \checkmark & \\
        
        \citet{zhou2024languageconditionedlearningroboticmanipulation} & \checkmark & \(\star\) & & & & \(\star\) & \\
        
        \rowcolor{gray!10}
        \citet{wang2024large} & \checkmark & & & & & \(\star\) & \\
        
        \citet{cao2024surveylargelanguagemodelenhanced} & \(\star\) & \checkmark & & & & & \\
        
        \rowcolor{gray!10}
        \citet{zhu2023diffusion} & \(\star\) & & & \checkmark & & & \\
        
        \textbf{Ours} & \(\star\) & \checkmark & \checkmark & \checkmark & \checkmark & \checkmark & \checkmark \\
        \bottomrule
    \end{tabular}
    \vspace{0.3cm}
    \captionsetup{justification=centering}
    \caption{Comparison of survey papers across robotics learning categories and methods: \textit{Foundation Models for Robotics} (\textbf{RFM}), \textit{Generative Tools for RL} (\textbf{Tools for RL}) and \textit{RL for Generative Policies} (\textbf{RL Policies}). \( \star \) means that the survey falls into that category but it does not broadly focus on the RL analysis, instead focusing on a specific subtopic.}
    \label{tab:comparison}
\end{table}

The main \textbf{contributions} of our work are:
{\small % or \footnotesize
\begin{enumerate}[noitemsep, topsep=0pt, leftmargin=*]
    \item A comprehensive review of the intersection between Transformer- and Diffusion-based generative models and RL for robotics.
    \item The first, to the best of our knowledge, dual analysis of how generative AI tools improve RL and RL improves generative policy models for robotics.
    \item The identification of best practices and challenges when using generative AI models as tools for RL.
    \item A detailed classification of RL-based training, fine-tuning and distillation methods for generalist generative policies.
    \item The identification of three new research directions integrating generative AI and RL to enhance robotics.
    \item A unified new taxonomy and continuously updated repository~\footnote{\url{https://github.com/clmoro/Robotics-RL-FMs-Integration}} for tracking progress in this field.
\end{enumerate}
}
The remainder of our paper is structured as follows: \textbf{Section~\ref{sec:taxonomy}} presents our taxonomy based on the duality between generative AI transformer- and diffusion-based tools and RL. \textbf{Section~\ref{sec:basemodel}} (\textit{Base Model}), \textbf{Section~\ref{sec:modality}} (\textit{Modality}) and \textbf{Section~\ref{sec:task}} (\textit{Task}) investigate deeper the dimension of generative AI models used as modular tools for the RL training loop, analyzing \textit{Generative Tools for RL}. On the other hand, \textbf{Section~\ref{sec:genpolicies}} explores \textit{RL-Based Training} of generative policies with consequent \textit{RL-Based Fine-Tuning} for downstream tasks (\textbf{Section~\ref{sec:finetuning}}) and \textit{Model Distillation} (\textbf{Section~\ref{sec:distillation}}), looking into \textit{RL for Generative Policies}. \textbf{Section~\ref{sec:challenges}} discusses challenges. We conclude with \textbf{Section~\ref{sec:conclusions}}, suggesting our perspective on possible future research directions based on our findings.

\subsection{Methodology}
We conducted this comprehensive review on the integration of transformer- and diffusion-based models with RL because these are two of the most rapidly emerging research areas in robotics learning. As noted in the previous section, we have observed a significant increase in the number of published papers on these topics in recent years, but especially, at their intersection. Given the exponential growth of research in generative AI for robotics, we chose to focus our review specifically on transformer- and diffusion-based models.
To compile our dataset, we searched for relevant papers published between 2019, when we identified the first preliminary works integrating transformers with RL for robotics applications, and February 2025. We used keyword-based searches to filter papers aligned with the scope of our review, excluding those that did not meet our criteria, following the widely recognized PRISMA (Preferred Reporting Items for Systematic Reviews and Meta-Analyses) methodology\footnote{\href{https://www.prisma-statement.org/}{PRISMA: Preferred Reporting Items for Systematic Reviews and Meta-Analyses}}. In the identification phase, we selected 15 relevant keywords to identify papers relevant to our review. We used combinations of terms such as RL with foundation models, LLM, VLM, diffusion models, world models, and video prediction models for filtering works on \textit{Generative Tools for RL}. For \textit{RL for Generative Policies}, we combined RL with transformer, diffusion, flow matching, generative policy, pre-training, fine-tuning, and distillation. Our search spanned multiple platforms, including Scopus, DBLP, IEEE Xplore, Google Scholar, and ArXiv, as well as references found in other related surveys listed in Table~\ref{tab:comparison}. In the screening phase, two researchers manually reviewed the abstracts of the identified papers to ensure their relevance to the robotics domain. During the selection stage, 169 papers that strictly met the inclusion criteria were chosen for analysis. To structure our taxonomy, we identified relevant categories based on two major dimensions of our study. Papers included in our review were required to be directly related to robotics and RL, published in English, and ideally peer-reviewed. However, given the fast-paced nature of this field, many relevant works have been published on preprint servers such as ArXiv and have yet to undergo formal peer review for top-tier journals or conferences.

\section{Taxonomy}
\label{sec:taxonomy}
\begin{figure}[!htbp]
      \centering
      \includegraphics[width=0.7\textwidth]{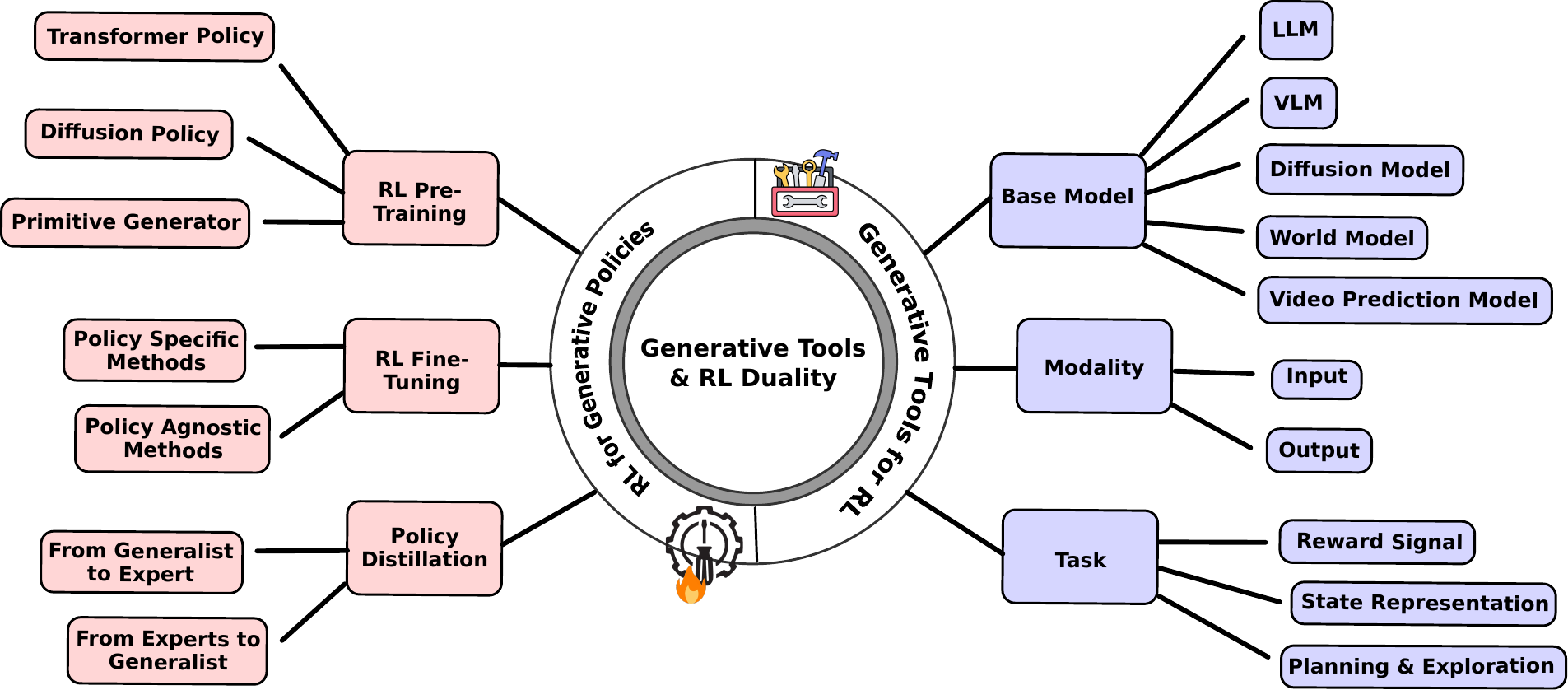}
      \captionsetup{justification=centering}
      \caption{\textbf{Taxonomy.} A taxonomy of the generative AI tools and RL integration for robotics.}
      \label{fig:taxonomy}
\end{figure}

In our review, we analyze how prominent generative AI tools and RL converge to enhance control policies for robots, specifically, action reference signals in the robot's Cartesian operational space. To achieve this, we propose a new taxonomy that categorizes the 169 papers we examined, tracing the evolution of generative AI and RL integration for multi-modal data fusion in robotics policy generation. Our taxonomy (see Figure~\ref{fig:taxonomy} for an abstract representation) primarily explores the duality between \textit{Generative AI tools for RL} and \textit{RL for Generative Policies}. We chose to analyze only papers focusing on specific generative models---particularly modern architectures widely adopted in recent works for policy generation, primarily based on transformers or diffusion models. We further classify the approaches along six principal dimensions based on orthogonal features of the works. For \textit{Generative Tools for RL}, we categorize papers based on their underlying model architecture, which we refer to as the \textbf{Base Model}; the input and output modalities, referred to as \textbf{Modality}; and finally, the aim of the RL process, which we call the \textbf{Task}. For \textit{RL for Generative Policies}, we analyze research along: pre-training strategies for different policy architectures, which we refer to as \textbf{RL pre-training}; methods for adapting policies through interaction, referred to as \textbf{RL fine-tuning}; and the transfer or compression of policies, which we call \textbf{Policy Distillation}.
We chose these categories to reflect two clear and growing trends in recent literature, that we depict in Figure~\ref{fig:duality}: the increasing use of generative models as tools within RL pipelines (blue outline), and the rise of RL techniques for training and refining generative policies (red outline), especially in robotic manipulation. Model architecture, modality, and task alignment are now central to foundation model choice~\citep{gui2024conformal}, while RL techniques like fine-tuning and policy distillation are increasingly used to adapt generative policies~\citep{mark2024policy}. Our taxonomy captures this duality by focusing on the most recurrent and emphasized dimensions in recent works.

\subsection{Generative Tools for RL} \textit{Generative Tools for RL} explores how selected generative AI architectures can be integrated into the RL training loop (see Figure~\ref{fig:subfig1}). We analyze prior work on leveraging generative and foundation models to enhance robotics, focusing on Transformer and Diffusion backbones.
\textbf{``As tools''} highlights that pre-trained foundation models (like LLMs) are not being retrained end-to-end with the RL agent, but are instead leveraged in a modular way---as plug-and-play components that provide capabilities (such as understanding or generating specific modalities) that the RL agent can use during training or decision making. Their in-context learning ability allows them to flexibly adapt to tasks without needing full retraining, making them useful tools for enhancing RL agents.
Our analysis is structured along three key secondary dimensions: (i) \textit{Base Model}, (ii) \textit{Modality}, and (iii) \textit{Task}.
\subsubsection*{Base Model} \textit{Base Model} classifies the models based on their underlying architecture, which directly influences their capabilities. Different architectures process and generate various types of data, shaping how they interact with RL. We analyze how deeply each base model can be incorporated into the RL framework based on its structural characteristics. Furthermore, for each model, approaches are analyzed along with (i) the framework names, (ii) code availability, (iii) the data used for training, and (iv) whether the pipeline is in simulation or the real world. Under this dimension, we consider the following models: \textit{LLMs}, \textit{VLMs}, \textit{Diffusion Models}, \textit{World Models}, and \textit{Video Prediction Models}.
\subsubsection*{Modality} Under the \textit{Modality} dimension, we analyze approaches based on their \textit{Input} and \textit{Output} modalities, such as text, images, trajectories, or low-level sensory signals. These modalities determine how generative AI integrates with RL by tackling input data interpretation and representation. Understanding these modalities helps assess the suitability of different generative AI tools for various RL applications.
\subsubsection*{Task} Generative AI models often serve as powerful priors within the RL training loop, as they are typically pre-trained modules that enhance specific aspects of learning. In this section, we delve deeper into how different works have employed generative models to address key \textit{Tasks} in the RL training loop such as: \textit{Reward Signal} generation, \textit{State Representation} and \textit{Planning \& Exploration}. Figure~\ref{fig:subfig1} illustrates scenarios where various generative AI tools enhance RL tasks. As an example, LLMs support symbolic reasoning for rewards or plan generation based on task descriptions, while VLMs contribute to reward augmentation or plan generation through scene understanding.

\subsection{RL for Generative Policies} 
\begin{figure}[!htbp]
    \centering
    \begin{subfigure}[t]{0.48\textwidth}
        \centering
        \includegraphics[width=\textwidth]{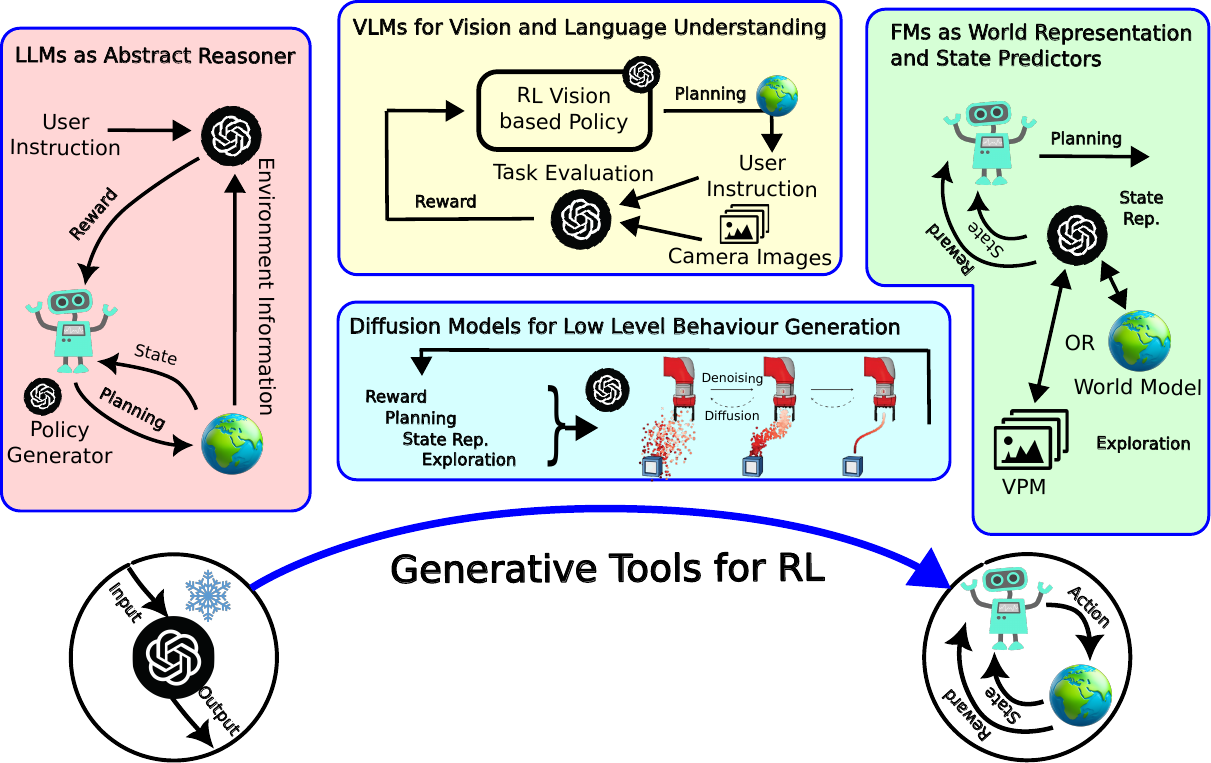}
        \captionsetup{justification=centering}
        \caption{This figure illustrates generative AI models (\textit{e.g.}, LLMs, VLMs, diffusion models, world/video prediction models) as tools for RL-based decision making.}
        \label{fig:subfig1}
    \end{subfigure}
    \hfill
    \begin{subfigure}[t]{0.48\textwidth}
        \centering
        \includegraphics[width=\textwidth]{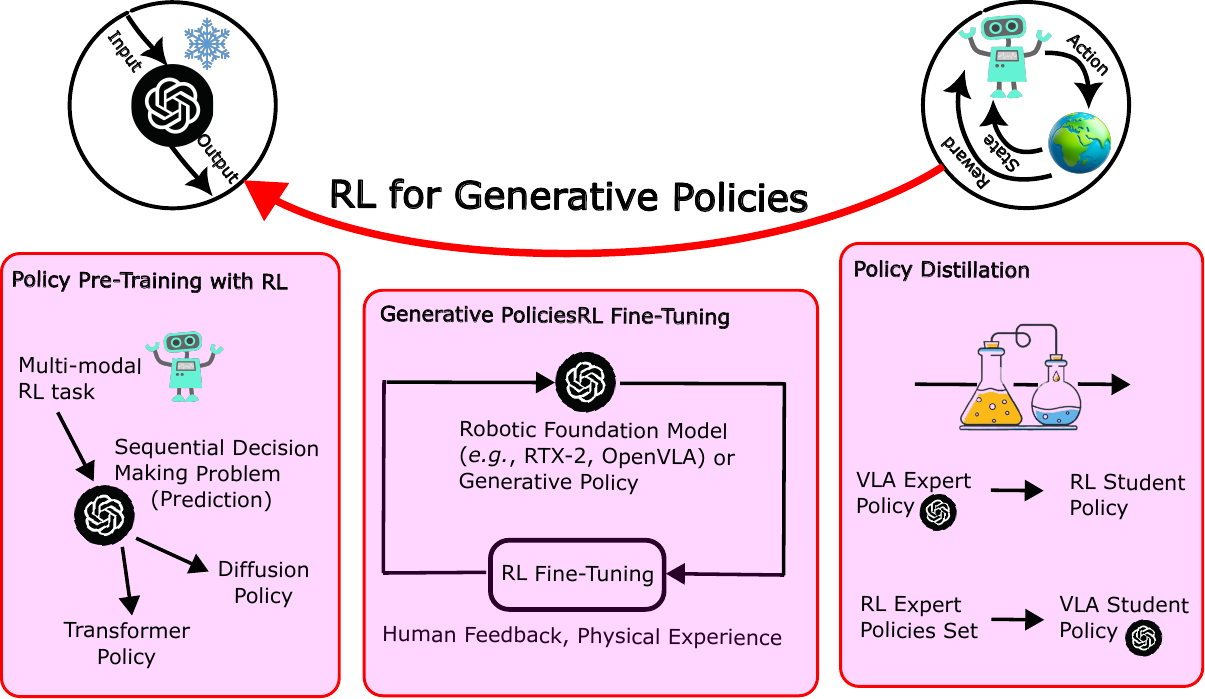}
        \captionsetup{justification=centering}
        \caption{This figure illustrates RL-based pre-training, fine-tuning and policy distillation, highlighting the complementary role of RL in enhancing generative policies.}
        \label{fig:subfig2}
    \end{subfigure}
    \captionsetup{justification=centering}
    \caption{\textbf{Duality between RL and generative AI models in robotics.}}
    \label{fig:duality}
\end{figure}

The second primary dimension of our taxonomy examines RL methods used to train generative models, offering a complementary perspective to \textit{Generative AI Tools for RL}. Here, we analyze works that employ RL-based approaches to pre-train, fine-tune, or distill generative policies---where RL is used directly to optimize models for action generation. We organize our discussion along three secondary dimensions, which we refer to as: (i) \textit{RL-Based Pre-Training}, (ii) \textit{RL-Based Fine-Tuning}, and (iii) \textit{Policy Distillation} (see Figure~\ref{fig:subfig2}).
\subsubsection*{RL-Based Pre-Training}
We survey various RL methods used to pre-train Transformer- and Diffusion-based policy backbones, enabling generalist generative policies that can process complex instructions; we divide them in \textit{Transformer Policy} and \textit{Diffusion Policy} respectively. Alternatively, RL can be used to train any set of simpler, task-specific policies for low-level control. These policies serve as a library of primitives (\textit{Primitive Generator}), which can then be orchestrated by a foundation model-based planner for more complex behaviors.
\subsubsection*{RL-Based Fine-Tuning} Fine-tuning pre-trained policies is a standard approach to adapting models to new tasks or datasets. We classify fine-tuning methods for generative policies based on their architecture and policy size---we call them \textit{Policy Specific Methods}. Additionally, we discuss emerging policy-agnostic fine-tuning techniques that aim to improve generalization across different generative architectures---called \textit{Policy Agnostic Methods}.
\subsubsection*{Policy Distillation} Beyond policy fine-tuning, RL includes the concept of \textit{Policy Distillation}, which refers to transferring knowledge and learned skills from a ``teacher'' policy to a ``student'' policy~\citep{rusu2015policy}. Recently, researchers have begun investigating how to distill knowledge from large pre-trained Vision-Language-Action (VLA) models into smaller and efficient RL-based expert policies (\textit{From Generalist to Expert}), as well as how to use RL pre-trained single-task policies to inject more knowledge into VLA models (\textit{From Experts to Generalist}).

\section{Base Model}
\label{sec:basemodel}
Generative models as tools for RL can be primarily classified by their \textit{Base Model}---\textit{i.e.}, the backbone architecture. In our review, we identify five main architectures that are typically used in RL: LLMs, VLMs, diffusion models, world models, and video prediction models. Figure~\ref{fig:tools} visually represents these models, along with reference papers \citep{ma2023eureka,adeniji2023language,Huang2023DiffusionReward,yang2023learning} for each category and their associated input/output modalities.

The \textit{Base Model} section classifies the papers according to their architecture, briefly describes their features, and summarizes key aspects in tables. These aspects are important when selecting a tool for the RL tasks defined in Section~\ref{sec:taxonomy}. The table columns represent the following:

{\small % or \footnotesize
\begin{itemize}[noitemsep, topsep=0pt, leftmargin=*]

    \item \textit{Paper}: Lists the corresponding research papers.
    \item \textit{Framework}: Specifies the generative AI framework used.
    \item \textit{Base Model/ Architecture}: Specifies the generative AI architecture used.
    \item \textit{Network}: Indicates the actual neural network model (\textit{e.g.}, GPT-3.5, GPT-4 for LLMs).
    \item \textit{Code Available}: Highlights whether the implementation is accessible.
    \item \textit{Training Data}: Lists the environments or datasets employed.
    \item \textit{Task}: Describes the key task or feature handled by the framework (\textit{e.g.}, reward function design, exploration).
    \item \textit{Simulation} or \textit{Real Exp}: Distinguishes between simulated environments and real-world experiments.
\end{itemize}
}
Table \ref{tab:llm_complete_summary_1} offers details on LLM-based works; Table \ref{tab:vlm_complete_summary} covers VLMs; Tables \ref{tab:diffusion_models_summary_1} and \ref{tab:diffusion_models_summary_2} summarize diffusion models; and Table \ref{tab:wm_summary} presents information on world models and video prediction models.

\subsection{LLM}
\label{sec:LLMs}
LLMs are changing how agents learn through RL. These models enable agents to acquire skills more effectively and flexibly by generating high-level plans, interpreting instructions, and providing structured feedback~\citep{ma2023eureka,ma2024dreureka,ma2024explorllm,bhat2024grounding,zhang2024motiongpt,sun2024prompt}. LLMs are incorporated into RL systems mostly because of their capacity to leverage vast amounts of existing information~\citep{yu2023language,cao2024surveylargelanguagemodelenhanced}. Because of this integrated knowledge, agents are less penalized by learning everything from scratch, which facilitates exploration and convergence on efficient behaviors. For example, GPT-4 is utilized in the EUREKA framework~\citep{ma2023eureka} to automatically generate code for RL tasks. Similarly, in open-ended textual settings with undefined learning objectives, LMA3~\citep{colas2023augmenting} uses GPT-3.5.
Adaptability is another benefit. LLMs can take zero-shot decisions through prompts or updated instructions, in contrast to classic RL techniques that frequently demand for big datasets and retraining when situations change\citep{ahn2022can}. But flexibility is not without its drawbacks. Because pre-trained models' reactions are limited by the data they were trained on, they may perform poorly in contexts that are novel or that change quickly. If not handled appropriately, this can result in biased behavior and false assumptions~\citep{firoozi2023foundationmodelsroboticsapplications}.
However, adding LLMs to RL also presents a number of difficulties. Interpretability is a major problem. It is challenging to comprehend or follow the logic behind the outputs of LLMs because they function mostly as ``black boxes.'' In safety-critical applications such as robotics, this lack of transparency becomes a significant concern~\citep{chu2023acceleratingRL,du2023guiding,zeng2022socratic}. Furthermore, there are real-world limitations due to the computing requirements of LLMs~\citep{dubey2024llama,wang2024large,zhu2023minigpt} and few works utilize open-source models.

Recent work combining LLMs with RL is summarized in Table \ref{tab:llm_complete_summary_1}, which covers the main frameworks, models, code availability, training data, and use cases in the RL training loop. The BOSS framework~\citep{zhang2023bootstrap}, for instance, makes use of GPT-3.5 to provide skill chaining across long-horizon activities, assisting agents in learning complicated behaviors through the extension and reuse of acquired abilities. FoMo~\citep{lubana2023fomo} increases agents' sensitivity to task context by using LLMs to dynamically scale reward signals based on visual inputs. The functions that the LLMs perform, including reward design, plan creation, or skill discovery, are represented by the \textit{Task} in the table. These task modules are integrated into larger RL pipelines. More details are given in Sections~\ref{sec:modality} and \ref{sec:task}.

\begin{table}[t!]
\centering
\tiny
\renewcommand{\arraystretch}{1.5} % Slightly reduced row height for better proportions
\resizebox{\textwidth}{!}{%
  {\fontsize{11}{13}\selectfont
    \begin{tabular}{p{3.8cm}p{2.8cm}p{3.2cm}p{3.2cm}p{2.2cm}p{3.2cm}p{2.8cm}p{3.2cm}}
    \toprule
    \textbf{Paper} & \textbf{Framework} & \textbf{Base Model/ Architecture} & \textbf{Network} & \textbf{Code} & \textbf{Training Data} & \textbf{Task} & \textbf{Simulation or Real-World Exp.} \\
    \midrule
    
    \rowcolor{gray!10}
    \citet{chu2023acceleratingRL} & Lafite-RL & LLM, World Model & GPT 3.5 and GPT 4 & N/A & Simulated robotic environment & Reward Signal & Simulation \\
    
    \citet{colas2023augmenting} & Language Model Augmented Autotelic Agent (LMA3) & LLM & ChatGPT (gpt-3.5-turbo-0301) & N/A & Text-based environment (CookingWorld) & Planning \& Exploration & Simulation \\
    
    \rowcolor{gray!10}
    \citet{zhang2023bootstrap} & Bootstrapping your Own Skills (BOSS) & LLM & GPT 3.5 & \href{https://github.com/clvrai/boss}{Yes} & Alfred Dataset, Demonstration data & Planning \& Exploration & Simulation + Real-World \\
    
    \citet{ahn2022can} & SayCan & LLM & GPT-3, FLAN and LAMBDA & \href{https://github.com/google-research/google-research/tree/master/saycan}{Yes} & Various real-world robotic tasks & Planning \& Exploration & Real-World \\
    
    \rowcolor{gray!10}
    \citet{ma2023eureka} & EUREKA & LLM & GPT 4 & \href{https://github.com/eureka-research/Eureka}{Yes} & Variety of robotic tasks & Reward Signal & Simulation (NVIDIA Isaac Gym) \\
    
    \citet{lubana2023fomo} & Foundation Models as Reward Functions (FoMo Rewards) & LLM & N/A & N/A & Visual observations from agent's trajectory & Reward Signal & Simulation \\
    
    \rowcolor{gray!10}
    \citet{Huang2023GroundedDG} & Grounded Decoding (GD) & LLM & InstructGPT and PALM & N/A & Various embodied tasks & Planning \& Exploration & Simulation \\
    
    \citet{carta2023grounding} & Grounded Language Models (GLAM) & LLM & N/A & \href{https://github.com/flowersteam/Grounding_LLMs_with_online_RL}{Yes} & Textual environment (BabyAI-Text) & State Representation & Simulation \\
    
    \rowcolor{gray!10}
    \citet{du2023guiding} & Exploring with LLMs (ELLM) & LLM & GPT-3, GPT-2 and InstructGPT & \href{https://github.com/google-research/google-research/tree/master/socraticmodels}{Yes} & Various environments (Crafter, Housekeep) & State Representation & Simulation \\

    \citet{colas2020language} & Intrinsic Motivations And Goal Invention for Exploration (IMAGINE) & LLM & N/A & \href{https://github.com/flowersteam/Imagine}{Yes} & Procedurally-generated scenes & Planning \& Exploration & Simulation \\
    
    \rowcolor{gray!10}
    \citet{hu2023language} & LLM Prior Policy with Regularized RL (instructRL) & LLM & GPT 3.5, GPT J & \href{https://github.com/hengyuan-hu/instruct-rl}{Yes} & Toy game and Hanabi benchmark & State Representation & Simulation \\
    
    \citet{yu2023language} & LLM-based Reward Translator & LLM, VLM & GPT-4 & \href{https://github.com/google-deepmind/language_to_reward_2023}{Yes} & Simulated quadruped robot, dexterous manipulator robot tasks & Reward Signal & Simulation \\
    
    \rowcolor{gray!10}
    \citet{dalal2024plan} & Plan-Seq-Learn (PSL) & LLM & GPT-4 & \href{https://github.com/mihdalal/planseqlearn}{Yes} & Challenging robotics tasks benchmarks & Planning \& Exploration & Simulation \\
    
    \citet{song2023self} & Self-Refined Large Language Model & LLM & GPT-4 & \href{https://github.com/zhehuazhou/LLM_Reward_Design}{Available Soon} & Various continuous robotic control tasks & Reward Signal & Simulation (NVIDIA Isaac Gym) \\
    
    \rowcolor{gray!10}
    \citet{xietext2reward} & Text2Reward Framework & LLM & GPT-4 & \href{https://github.com/xlang-ai/text2reward} {Yes} & ManiSkill2, MetaWorld, MuJoCo & Reward Signal & Simulation \\

    \rowcolor{gray!10}
    \citet{triantafyllidis2024intrinsic} & Intrinsically Guided Exploration from Large Language Models (IGE-LLMs) & LLM & GPT-4 & N/A & Robotic manipulation tasks & Planning \& Exploration & Simulation \\
    
    \bottomrule
    \end{tabular}
  }
}
\vspace{0.3cm} % Reduced space above caption
\captionsetup{justification=centering}
\caption{Summary of large language models as tools for RL.}
\label{tab:llm_complete_summary_1}
\end{table}

\subsection{VLM}
\label{sec:VLMs}
The integration of multi-modal VLMs into RL is a significant step forward with respect to single modality LLMs, enabling more useful and informative text and visual input fusion~\citep{cui2022can,venuto2024code,adeniji2023language,ma2023liv,wang2024,sontakke2024roboclip,yang2024robot,di2023towards,rocamonde2023vision,baumli2023vision,chen2024vision,mahmoudieh2022zero,ma2024explorllm}. This is particularly useful in complex robotic tasks where goal states are better represented through images rather than numbers or text.
One of the key advantages of VLMs is their ability to generalize across tasks without requiring environment-specific fine-tuning, similar to LLMs~\citep{cui2022can}. Moreover, VLMs can work directly with real-world perception and don’t need to rely on accurate environment information translated into text, compared to LLM frameworks such as EUREKA~\citep{ma2023eureka}.
The computational demand of VLMs remains significant, if not greater than that of LLMs. Few methods~\citep{venuto2024code,wang2024} aim to mitigate this by solving RL tasks without excessive computational costs or calling the model inference at lower frequency. Moreover, fine-tuning VLMs to optimize their responses for RL specific tasks remains sometimes necessary~\citep{huang2024dark}.
The effectiveness of VLMs in RL also depends on the choice of architecture---\textit{e.g.}, several works use specialized modules such as CLIP (Contrastive Language-Image Pre-Training), a model that predicts the most relevant text sequence given an image~\citep{radford2021learning,shridhar2022cliport,sontakke2024roboclip}, while others use general-purpose foundation models like GPT-4---and experimental setup, inheriting the issues of LLMs. Some studies focus on simulation environments with synthetic data~\citep{wang2024}, while others explore real-world robotic applications~\citep{ma2024explorllm}, leading to variations in prompting strategies, RL methodologies, and evaluation metrics. Notably, in Table~\ref{tab:vlm_complete_summary} some models are used primarily for reward generation---similar to LLMs but based on visual input (\textit{e.g.}, \textit{Reward Generation} in \citet{venuto2024code})---while others use VLMs to define the goal task or condition the learning to it (classified as \textit{State Representation}), as seen in \citet{cui2022can}. See Section~\ref{sec:task} for more details on column \textit{Task}.

\begin{figure}[!htbp]
    \centering
    \begin{subfigure}[t]{0.48\textwidth}
        \centering
        \includegraphics[width=\textwidth]{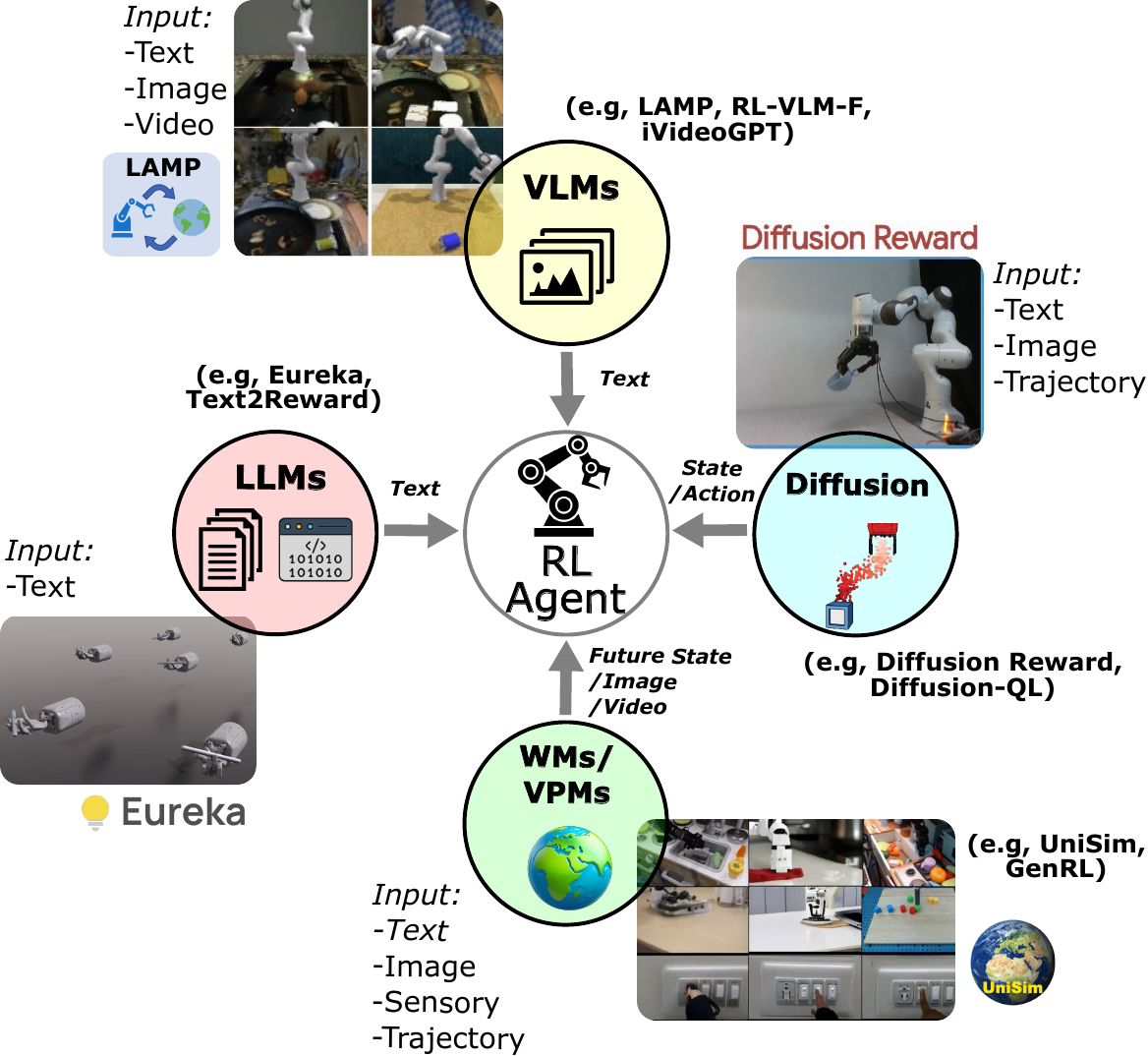}
        \captionsetup{justification=centering}
        \caption{Base models modularity.}
        \label{fig:tools}
        \end{subfigure}
    \hfill
    \begin{subfigure}[t]{0.48\textwidth}
        \centering
        \includegraphics[width=\textwidth, height=0.3\textheight]{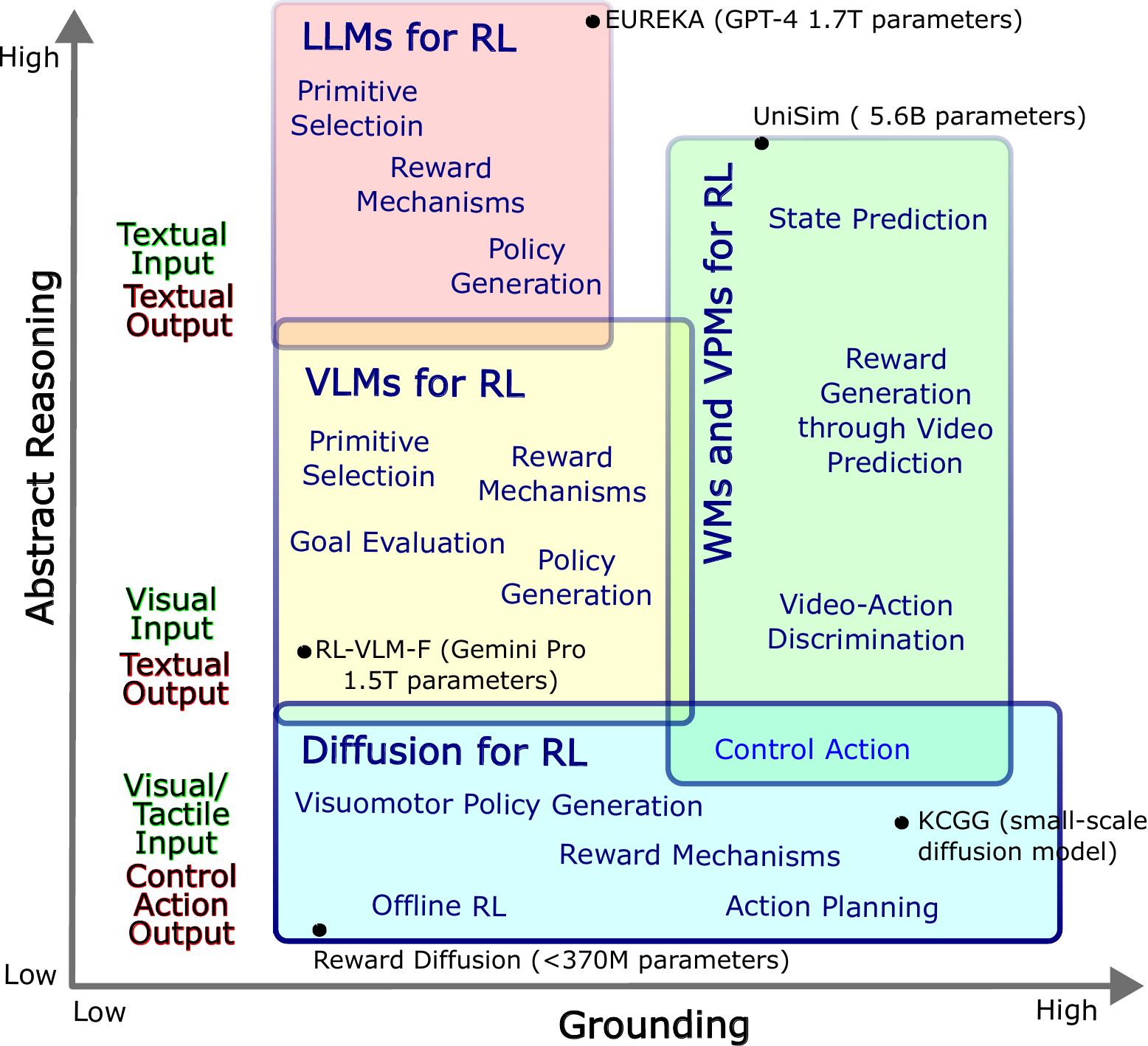}
        \captionsetup{justification=centering}
        \caption{Trade-off between abstraction and grounding, when using generative AI tools for RL.}
        \label{fig:grounding}
    \end{subfigure}
    \captionsetup{justification=centering}
    \caption{\textbf{Generative AI tools for RL.}}
\end{figure}

\begin{table}[t!]
\centering
\tiny
\renewcommand{\arraystretch}{1.4} % Adjusted row height
\resizebox{\textwidth}{!}{%
  {\fontsize{11}{13}\selectfont
  \begin{tabular}{p{3.6cm}p{2.8cm}p{2.8cm}p{3.2cm}p{2.2cm}p{3.2cm}p{2.8cm}p{3.2cm}}
  \toprule
  \textbf{Paper} & \textbf{Framework} & \textbf{Base Model/ Architecture} & \textbf{Network} & \textbf{Code} & \textbf{Training Data} & \textbf{Task} & \textbf{Simulation or Real-World Exp.} \\
  \midrule
  
  \rowcolor{gray!10}
  \citet{cui2022can} & Zero-Shot Task Specification (ZeST) & VLM & ImageNet-supervised ResNet50, ImageNet-trained MoCo, CLIP & N/A & Simulated robot manipulation tasks and real-world datasets & State Representation & Simulation \\
  
  \citet{venuto2024code} & VLM-CaR (Code as Reward) & VLM & GPT-4 & N/A & Discrete and continuous environments & Reward Signal & Simulation \\
  
  \rowcolor{gray!10}
  \citet{adeniji2023language} & Language Reward Modulated Pretraining (LAMP) & VLM, World Model & CLIP & \href{https://github.com/ademiadeniji/lamp}{Yes} & Ego4D, video Dataset, automatically Generated language Dataset & Reward Signal & Simulation \\
  
  \citet{ma2023liv} & Language-Image Value Learning (LIV) & VLM & CLIP with ResNet 50 backbone & \href{https://github.com/penn-pal-lab/LIV}{Yes} & EpicKitchen, simulated and real-world robot environments & Reward Signal & Simulation + Real-World \\
  
  \rowcolor{gray!10}
  \citet{wang2024} & RL-VLM-F & VLM & Gemini and GPT-4 Vision & \href{https://github.com/yufeiwang63/RL-VLM-F}{Yes} & Various domains including classic control and manipulation tasks & Reward Signal & Simulation \\
  
  \citet{sontakke2024roboclip} & RoboCLIP & VLM & S3D & \href{https://github.com/sumedh7/RoboCLIP/tree/main}{Yes} & Howto100M & Reward Signal & Simulation \\
  
  \rowcolor{gray!10}
  \citet{yang2024robot} & ROBOFUME & VLM, LLM & MiniGPT4 & \href{https://robofume.github.io/}{Available soon} & Diverse datasets, real robot experiments & Reward Signal & Simulation + Real-World \\
  
  \citet{di2023towards} & Unified Agent with Foundation Models & VLM, LLM & CLIP, FLAN-T5 & N/A & Image and Text datasets & State Representation & Simulation \\
  
  \rowcolor{gray!10}
  \citet{rocamonde2023vision} & Zero-Shot Vision-Language Models (VLM-RMs) & VLM & CLIP & \href{https://github.com/AlignmentResearch/vlmrm}{Yes} & Human motion task dataset & Reward Signal & Simulation \\
  
  \citet{baumli2023vision} & Vision-Language Models (VLMs) like CLIP & VLM & CLIP & N/A & Playhouse & Reward Signal & Simulation \\
  
  \rowcolor{gray!10}
  \citet{chen2024vision} & PR2L (Promptable Representations for RL) & VLM, LLM & Vicuna-7B version of the InstructBLIP, a Llama2-7B Prismatic VLM & \href{https://github.com/pr2l/pr2l.github.io/blob/master/static/notebooks/PR2LExample.ipynb}{Yes} & Minecraft, Habitat environments & State Representation & Simulation \\
  
  \citet{mahmoudieh2022zero} & Zero-Shot Reward Model (ZSRM) & VLM & ResNet-50 & N/A & Large dataset of captioned images & Reward Signal & Simulation \\
  
  \rowcolor{gray!10}
  \citet{ma2024explorllm} & ExploRLLM & VLM, LLM & GPT-4 & N/A & Image-based robot manipulation tasks dataset & Planning \& Exploration & Simulation + Real-World \\
  
  \bottomrule
  \end{tabular}
  }
}
\vspace{0.3cm} % Adjusted space above caption
\captionsetup{justification=centering}
\caption{Summary of vision-language models as tools for RL.}
\label{tab:vlm_complete_summary}
\end{table}

\subsection{Diffusion Model}
\label{sec:Diff}
Diffusion models, known for success in image generation~\citep{rombach2022high,esser2024scaling}, are emerging in RL~\citep{lu2023contrastive,kang2024efficient,suh2023fighting,hansen2023idql,kim2024robust,venkatraman2023reasoning,hu2023instructed,psenka2023learning,jain2023learning,chen2022offline}. Unlike models that use only textual or visual inputs, diffusion models can also operate directly in the trajectory space, generating continuous actions~\citep{zhu2023madiff,ni2023metadiffuser}.
These models learn to reverse a diffusion process that adds noise to states, rewards or policy parameters, generating new quantities from random noise. This approach has several advantages. First, it promotes behavior diversity, encouraging exploration and the discovery of novel solutions by generating a wide range of control policies. Second, it is well-suited for tasks with continuous action spaces, thereby avoiding the suboptimal performance that can arise from discretization. Finally, it improves sample efficiency by learning from limited demonstrations, reducing the need for extensive data collection~\citep{he2024diffusion,zhu2023diffusion,wang2022diffusion}.
The ability of diffusion models to be conditioned on language instructions or goal prompts enhances adaptability, making zero-shot policy generation possible. While still early in development, diffusion models offer new perspectives compared to more widely explored LLMs and VLMs~\citep{xiao2025safediffuser,liang2023adaptdiffuser,ni2023metadiffuser}. Their iterative process allows for fine adjustments, making them ideal for complex tasks like grasping and balancing~\citep{lee2024refining}. Additionally, their capacity to model complex dynamics and manage noise and uncertainty makes them robust in real-world robotic environments. However, their effectiveness in high-dimensional tasks is limited~\citep{he2024diffusion,zhou2024adaptive,liu2023dipper} and issues such as sensitivity to training data persist.
We investigate further these aspects in Section~\ref{sec:task} and we classify diffusion-based papers in Tables~\ref{tab:diffusion_models_summary_1} and \ref{tab:diffusion_models_summary_2}.

\begin{table}[t!]
\centering
\tiny
\renewcommand{\arraystretch}{1.4} % Adjusted row height
\resizebox{\textwidth}{!}{%
  {\fontsize{11}{13}\selectfont
  \begin{tabular}{p{4.2cm}p{2.8cm}p{2.3cm}p{1.0cm}p{5.3cm}p{2.8cm}p{3.5cm}}
  \toprule
  \textbf{Paper} & \textbf{Framework} & \textbf{Base Model/ Architecture} & \textbf{Code} & \textbf{Training Data} & \textbf{Task} & \textbf{Simulation or Real-World Exp.} \\
  \midrule
  
  \rowcolor{gray!10}
  \citet{lu2023contrastive} & Contrastive Energy Prediction (CEP) & Diffusion model & \href{https://github.com/ChenDRAG/CEP-energy-guided-diffusion}{Yes} & D4RL benchmarks & State Representation & Simulation (D4RL benchmarks) \\
  
  \citet{kang2024efficient} & EDP & Diffusion model & \href{https://github.com/sail-sg/edp}{Yes} & D4RL benchmark datasets & State Representation & Simulation (D4RL benchmark) \\
  
  \rowcolor{gray!10}
  \citet{suh2023fighting} & Score-Guided Planning (SGP) & Diffusion model & \href{https://github.com/hjsuh94/score_po}{Yes} & Cart-pole system, D4RL benchmark (MuJoCo tasks), pixel-based single integrator environment & Planning \& Exploration & Simulation (Various simulations) \\
  
  \citet{hansen2023idql} & Implicit Diffusion Q-Learning (IDQL) & Diffusion model & \href{https://github.com/philippe-eecs/IDQL}{Yes} & D4RL benchmark (halfcheetah, hopper, walker2d, antmaze), Maze2D & State Representation & Simulation (D4RL benchmark, Maze2D) \\
  
  \rowcolor{gray!10}
  \citet{kim2024robust} & DuSkill & Diffusion Model & N/A & Rule-based expert policies, multi-stage Meta-World tasks (slide puck, close drawer) & State Representation & Simulation (Multi-stage Meta-World) \\
  
  \citet{venkatraman2023reasoning} & LDCQ & Diffusion Model, World Model & N/A & MuJoCo benchmarks (Hopper, Walker, Ant), Maze2D, AntMaze, FrankaKitchen, CARLA & State Representation & Simulation (Various environments) \\
  
  \rowcolor{gray!10}
  \citet{hu2023instructed} & Temporally-Composable Diffuser (TCD) & Diffusion Model & N/A & Gym-MuJoCo environments (HalfCheetah, Hopper, Walker2D), Maze2D, Hand Manipulation tasks & State Representation & Simulation (Gym-MuJoCo, Maze2D, Hand Manipulation) \\
  
  \citet{psenka2023learning} & Q-Score Matching (QSM) & Diffusion Model & \href{https://github.com/Alescontrela/score_matching_rl}{Yes} & DeepMind Control Suite (Cartpole Balance, Cartpole Swingup, Cheetah Run, Hopper Hop, Walker Walk, Walker Run, Quadruped Walk, Humanoid Walk) & State Representation & Simulation (DeepMind Control Suite) \\
  
  \rowcolor{gray!10}
  \citet{jain2023learning} & Merlin & Diffusion Model & N/A & PointReach, PointRooms, Reacher, SawyerReach, SawyerDoor, FetchReach, FetchPush, FetchPick, FetchSlide, HandReach & State Representation & Simulation (Various simulated environments) \\
  
  \citet{zhu2023madiff} & MADIFF & Diffusion Model & \href{https://github.com/zbzhu99/madiff}{Yes} & Multi-agent particle environments, MA Mujoco, StarCraft Multi-Agent Challenge (SMAC), NBA dataset & State Representation & Simulation (Multi-agent environments) \\
  
  \rowcolor{gray!10}
  \citet{ni2023metadiffuser} & MetaDiffuser & Diffusion Model & \href{https://metadiffuser.github.io/}{Yes} & MuJoCo benchmarks (Hopper-Param, Walker-Param), Point-Robot 2D navigation & State Representation & Simulation (MuJoCo, Point-Robot) \\
  
  \citet{chen2022offline} & SfBC & Diffusion Model & N/A & D4RL benchmarks, AntMaze tasks, Maze2d, FrankaKitchen, Bidirectional-Car tasks & State Representation & Simulation (Various benchmarks) \\
  
  \bottomrule
  \end{tabular}
  }
}
\vspace{0.3cm} % Adjusted space above caption
\captionsetup{justification=centering}
\caption{Summary of diffusion models as tools for RL.}
\label{tab:diffusion_models_summary_1}
\end{table}

\begin{table}[t!]
\centering
\tiny
\renewcommand{\arraystretch}{1.4} % Adjusted row height
\resizebox{\textwidth}{!}{%
  {\fontsize{11}{13}\selectfont
  \begin{tabular}{p{4.2cm}p{2.8cm}p{2.3cm}p{1.0cm}p{4.3cm}p{3.3cm}p{2.8cm}}
  \toprule
  \textbf{Paper} & \textbf{Framework} & \textbf{Base Model/ Architecture} & \textbf{Code} & \textbf{Training Data} & \textbf{Task} & \textbf{Simulation or Real-World Exp.} \\
  \midrule
  
  \rowcolor{gray!10}
  \citet{liang2023adaptdiffuser} & AdaptDiffuser & Diffusion Model & \href{https://github.com/Liang-ZX/adaptdiffuser}{Yes} & Synthetic expert data & Planning \& Exploration & Simulation (Maze2D, MuJoCo) \\
  
  \citet{he2024diffusion} & Multi-Task Diffusion Model (MTDIFF) & Diffusion model & N/A & Meta-World and Maze2D environments & Planning \& Exploration & Simulation (Meta-World, Maze2D) \\
  
  \rowcolor{gray!10}
  \citet{brehmer2024edgi} & EDGI & Diffusion Model, World Model & N/A & Offline trajectory datasets (3D navigation, Kuka robotic arm) & Planning \& Exploration & Simulation \\
  
  \citet{li2023hierarchical} & Hierarchical Diffusion (HDMI) & Diffusion model, World Model & N/A & Maze2D, AntMaze, D4RL environments, NeoRL benchmark & Planning \& Exploration & Simulation (Various environments) \\
  
  \rowcolor{gray!10}
  \citet{xiao2023safediffuser} & SafeDiffuser & Diffusion model & \href{https://github.com/Weixy21/SafeDiffuser}{Yes} & Maze2D environments, MuJoCo environments (Walker2D, Hopper), Pybullet environments & Planning \& Exploration & Simulation (Multiple environments) \\
  
  \citet{chen2024simple} & Hierarchical Diffuser & Diffusion Model, World Model & N/A & Maze2D (U-Maze, Medium, Large), Multi2D, AntMaze, Gym-MuJoCo, FrankaKitchen & Planning \& Exploration & Simulation (Various benchmarks) \\
  
  \rowcolor{gray!10}
  \citet{kim2024stitching} & Sub-trajectory Stitching with Diffusion (SSD) & Diffusion Model & \href{https://github.com/rlatjddbs/SSD}{Yes} & Maze2D environments, Fetch environments & Planning \& Exploration & Simulation (Maze2D, Fetch) \\
  
  \citet{lee2024refining} & Restoration Gap Guidance (RGG) & Diffusion model & \href{https://github.com/leekwoon/rgg}{Yes} & Maze2D environments, Gym-MuJoCo locomotion tasks, block stacking tasks with Kuka iiwa robotic arm & Planning \& Exploration & Simulation (Multiple environments) \\
  
  \rowcolor{gray!10}
  \citet{zhang2024language} & Language Control Diffusion (LCD) & Diffusion Model & \href{https://github.com/ezhang7423/language-control-diffusion}{Yes} & CALVIN language robotics benchmark, CLEVR-Robot benchmark & Planning \& Exploration & Simulation (CALVIN, CLEVR-Robot) \\
  
  \citet{janner2022diffuser} & Diffuser & Diffusion Model & \href{https://github.com/jannerm/diffuser}{Yes} & Maze2D environments, block stacking tasks, D4RL locomotion suite & Planning \& Exploration & Simulation (Multiple environments) \\
  
  \rowcolor{gray!10}
  \citet{nuti2024extracting} & Relative Reward Function & Diffusion model & \href{https://github.com/FelipeNuti/diffusion-relative-rewards}{Yes} & Maze2D, D4RL locomotion tasks, I2P dataset & Reward Signal & Simulation (Maze2D, D4RL) \\
  
  \citet{mazoure2023value} & Diffused Value Function (DVF) & Diffusion Model, World Model & N/A & Maze2D environments, PyBullet environments, D4RL offline suite & Reward Signal & Simulation (Multiple environments) \\
  
  \bottomrule
  \end{tabular}
  }
}
\vspace{0.3cm} % Adjusted space above caption
\captionsetup{justification=centering}
\caption{Summary of diffusion models as tools for RL (continued).}
\label{tab:diffusion_models_summary_2}
\end{table}

\subsection{World Model and Video Prediction Model}
\label{sec:WM}
World models in RL introduce new methods for learning representations and dynamic models that serve as internal simulators for planning and prediction in RL~\citep{wu2023daydreamer,mazzaglia2024multimodal,hassan2024gem,Zala2024EnvGen}. Typically, a world model consists of an encoder that incorporates environmental observations, a dynamics model that is learned during pre-training and allows for internal simulation, and an optional decoder that reconstructs information from the latent space. A reward model, which forecasts rewards based on the learned representation, might also be included. Although they are not commonly employed as primary dynamic predictors, LLMs and VLMs have recently being integrated into world models. LLMs are mainly used for high-level reasoning and task specification~\citep{Wang2023GenSimGR,nottingham2023embodied}, while VLMs~\citep{mazzaglia2024multimodal} support state representation by providing rich visual-linguistic embeddings. The core transition dynamics are typically modeled using architectures like RNNs, transformers, or diffusion models.

Table~\ref{tab:wm_summary} demonstrates that \textit{State Representation} is the primary use of world models, with almost all works concentrating on this function~\citep{ha2018recurrent,mazzaglia2024multimodal,wu2024ivideogpt,wang2023robogen}. Few studies focus on \textit{Reward Signal} generation~\citep{nottingham2023embodied,chen2021learning}, or \textit{Planning \& Exploration}~\citep{yang2023learning,ye2023foundation} as the main tasks. With little application to real-world data, the majority of the studied techniques are trained and assessed in simulated contexts. Though, few works~\citep{Wang2023GenSimGR,seo2023multi,ye2023foundation,chen2021learning}, include real-world experiments.

Video prediction models, special models that learn the temporal dynamics of a visual environment, have also proved good results in visual RL, with frameworks like VIPER~\citep{escontrela2024video}. However, the main limitation is the reliance on specific environments for training~\citep{chen2021learning,alakuijala2023learning}.

See Table~\ref{tab:vp_vs_wm} for a comparison between features of VPMs and WMs for RL, where \textit{action-conditioned} means that the model's prediction can be influenced also by the action taken by the RL agent, and Sections~\ref{sec:task} for more details.

\begin{table}[!htbp]
\centering
\tiny
\begin{tabular}{lp{4.1cm}p{4.2cm}}
\toprule
\textbf{Feature} & \textbf{Video Prediction Models} & \textbf{World Models} \\
\midrule

\rowcolor{gray!10}
\textbf{Output} & 
\makecell[l]{Future frames (pixels)} & 
\makecell[l]{Latent states, frames, rewards} \\

\rowcolor{white}
\textbf{Use Case} & 
\makecell[l]{Future frame generation} & 
\makecell[l]{Future state/control generation} \\

\rowcolor{gray!10}
\textbf{Reward Modeling} & 
\makecell[l]{Rarely included} & 
\makecell[l]{Often included} \\

\rowcolor{white}
\textbf{Action-Conditioned} & 
\makecell[l]{No} & 
\makecell[l]{Usually} \\

\bottomrule
\end{tabular}
\captionsetup{justification=centering}
\caption{\textbf{Comparison between video prediction models and world models for RL.}}
\label{tab:vp_vs_wm}
\end{table}

\begin{table}[t!]
\centering
\tiny
\renewcommand{\arraystretch}{1.4} % Adjusted row height
\resizebox{\textwidth}{!}{%
  {\fontsize{11}{13}\selectfont
  \begin{tabular}{p{4.2cm}p{2.8cm}p{2.3cm}p{1.0cm}p{4.3cm}p{3.3cm}p{2.8cm}}
  \toprule
  \textbf{Paper} & \textbf{Framework} & \textbf{Model Class} & \textbf{Code} & \textbf{Training Data} & \textbf{Task} & \textbf{Simulation or Real-World Exp.} \\
  \midrule
  
  \rowcolor{gray!10}
  \citet{ha2018recurrent} & MDN-RNN & World Model & \href{https://blog.otoro.net/2018/06/09/world-models-experiments/}{Yes} & CarRacing-v0, DoomTakeCover-v0 & State Representation & Simulation \\
  
  \citet{Wang2023GenSimGR} & GENSIM (GPT-4, GPT-3.5, Code Llama) & World Model, LLM & \href{https://github.com/liruiw/GenSim}{Yes} & Generated 100 tasks from a task library & State Representation & Simulation + Real-World \\
  
  \rowcolor{gray!10}
  \citet{mazzaglia2024multimodal} & MFWM (GRU-based, InternVideo2) & World Model, LLM, VLM & \href{https://github.com/mazpie/genrl}{Yes} & Walker, Cheetah, Quadruped, Stickman, Kitchen & State Representation & Simulation \\
  
  \citet{wu2024ivideogpt} & iVideoGPT & World Model & \href{https://github.com/thuml/iVideoGPT}{Yes} & OXE Dataset, SSv2, RoboNet & State Representation & Simulation \\
  
  \rowcolor{gray!10}
  \citet{mao2024zero} & Segment Anything Model (SAM), LLMs (GPT-3.5) & World Model, LLM & N/A & Cart Pole, Lunar Lander & State Representation & Simulation \\
  
  \citet{bruce2024genie} & Genie (ST-Transformer) & World Model & N/A & Platformers dataset, robotics dataset & State Representation & Simulation \\
  
  \rowcolor{gray!10}
  \citet{yang2023learning} & Observation Prediction Model (Video Diffusion Model) & World Model, VLM, Diffusion Model & N/A & Simulated environments, real robot data, human activity videos, panorama scans & Planning \& Exploration & Simulation \\
  
  \citet{seo2023masked} & Masked World Models (MWM) & World Model & \href{https://github.com/younggyoseo/MWM}{Yes} & Meta-world, RLBench & State Representation & Simulation \\
  
  \rowcolor{gray!10}
  \citet{seo2023multi} & Multi-View Masked World Models (MV-MWM) & World Model & \href{https://github.com/younggyoseo/MV-MWM}{Yes} & RLBench & State Representation & Simulation + Real-World \\
  
  \citet{ye2023foundation} & Foundation Actor-Critic (FAC) & World Model, Diffusion Model & N/A & Internet-scale robotics datasets, Meta-World & Planning \& Exploration & Simulation + Real-World \\
  
  \rowcolor{gray!10}
  \citet{chen2021learning} & Domain-agnostic Video Discriminator (DVD) & World Model & N/A & Something-Something-V2, robot videos in various environments & Reward Signal & Simulation + Real-World \\
  
  \citet{nottingham2023embodied} & DECKARD & World Model, LLM & \href{https://github.com/DeckardAgent/deckard}{Yes} & Minecraft environment & Reward Signal & Simulation \\
  
  \rowcolor{gray!10}
  \citet{Zala2024EnvGen} & EnvGen & World Model, LLM & \href{https://github.com/aszala/envgen}{Yes} & Generated and original environments & State Representation & Simulation \\
  
  \citet{wang2023robogen} & RoboGen Generative Simulation & World Model, LLM & \href{https://github.com/Genesis-Embodied-AI/RoboGen}{Yes} & Generated tasks, scenes & State Representation & Simulation \\
  
  \bottomrule
  \end{tabular}
  }
}
\vspace{0.3cm} % Adjusted space above caption
\captionsetup{justification=centering}
\caption{Summary of world models as tools for RL.}
\label{tab:wm_summary}
\end{table}

\section{Modality}
\label{sec:modality}
This section focuses on the classification of five types of generative AI models used in RL, as introduced in Section~\ref{sec:basemodel}, with an emphasis on how their input/output modalities shape their role within RL frameworks.

Bridging the gap between symbolic concepts and real-world experience is a concept referred to as \textit{grounding}, introduced in Section~\ref{sec:introduction}. Our classification highlights a fundamental trade-off between abstraction and grounding: while LLMs~\citep{dubey2024llama,achiam2023gpt} and VLMs~\citep{alayrac2022flamingo,zhang2024vision,bordes2024introduction,zhu2023minigpt,chen2024vision} are adept at abstract reasoning and symbolic processing, their input/output operations are less directly tied to physical control~\citep{carta2023grounding,bhat2024grounding,cohen2024survey,ahn2022can,ma2023eureka,xietext2reward}. On the other hand, diffusion models, by generating low-level, continuous control actions and operating directly in the action space, provide a more sample-efficient approach to policy learning~\citep{ho2022classifier,rombach2022high,wang2022diffusion,chi2023diffusion,zhu2023diffusion,janner2022diffuser}, though they may lack the higher-level abstraction capabilities inherent to larger language models. Models that can effectively fuse abstraction and grounding for low-level action generation in robotics remain limited, although world models~\citep{wu2023daydreamer,seo2023masked,yang2023learning} and video prediction models~\citep{escontrela2024video,du2023learning} represent promising approaches by integrating both high-level abstraction and grounded sensory information.

The diagram in Figure~\ref{fig:grounding} shows that, based on our analysis, each generative AI tool used in RL contributes distinct input/output modalities. LLMs, working with text as both input and output excel at symbolic reasoning and can synthesize or interpret textual inputs such as task goals, environment documentation, or learned RL primitives, enabling them to produce reward signals or task refinements aligned with high-level objectives~\citep{ma2023eureka,ma2024dreureka}. VLMs, in contrast, can process visual inputs and generate reasoning over visual scenes. This makes them valuable for visual feedback mechanisms~\citep{wang2024}. Their ability to bridge visual and textual modalities allows flexible integration into tasks where visual context is critical. Diffusion models can be highly specialized for policy learning and state generation, as they can model complex distributions over low-level states and actions. Operating directly in continuous action spaces, they can produce precise control signals that are easily integrated with RL algorithms~\citep{chi2023diffusion,Huang2023DiffusionReward}. This modality specificity makes them ideal for applications in robotics and control where fine-grained action generation is required. Lastly, world and video prediction models, with their broad support for multi-modal input fusion and internal representation learning capacity, provide an even more flexible foundation. They can incorporate and generate rich multi-modal state representations (including visual, textual, proprioceptive, and other sensory data), making them well-suited for learning predictive models of environment dynamics and for supporting planning and model-based RL~\citep{wu2023daydreamer,seo2023masked,yang2023learning}.

The level of abstraction and the size of the models (in terms of parameters) strongly influence how easily they can be integrated into the RL training loop. Larger, more powerful, and generalist models (\textit{e.g.}, current LLMs, which are text-based) can be quickly incorporated in a zero-shot fashion in several frameworks~\citep{ma2023eureka,ma2024dreureka}. However, they are often less tailored and less adaptable to the specific input/output requirements of RL tasks. Moreover, the choice of model---whether a diffusion model, an LLM, or a VLM---directly impacts the integration strategy within RL frameworks. Different models impose distinct computational demands and offer varying levels of flexibility in adapting to real-world dynamics~\citep{he2024diffusion,zhou2024diffusion,Huang2023DiffusionReward}.
The diagram in Figure~\ref{fig:grounding} visually summarizes these trade-offs across models, mapping representative tools based on their degree of symbolic reasoning and grounding in physical control. This complements our analysis of modality diversity and integration strategies by illustrating how models such as \textbf{UniSim}, \textbf{EUREKA}, \textbf{RL-VLM-F}, \textbf{Reward Diffusion}, and \textbf{KCGG} occupy different positions in this space.

\section{Task}
\label{sec:task}
RL tries to perform optimal decision-making considering the interaction of an agent (\textit{e.g.}, a robot) with its environment, with the goal of maximizing rewards \citep{sutton1988learning,beck2023survey}. In the RL framework, several core components play essential roles in guiding the agent's behavior. First, we have \textit{states}, which represent all possible situations the agent might encounter within its environment. At any given moment, the agent will find itself in a particular state, prompting it to consider the best \textit{action} (\textit{e.g.}, the action that maximizes the reward). These actions are the set of choices available to the agent, allowing it to interact with and influence the environment. The \textit{policy} represents the probability of taking action given a state (\textit{e.g.}, it's the agent’s strategy). As the agent takes actions, it receives feedback from the environment through the \textit{reward function}, which assigns immediate rewards based on the outcome of each action. This reward signal helps the agent understand how to act optimally to achieve long-term goals.
Two key functions support the agent in making more informed decisions over time. The \textit{value function} estimates the expected long-term return of being in a particular state, giving the agent an idea of how promising that state is under the current policy. The \textit{Q-function}, on the other hand, is slightly more detailed. It assesses the expected return not just for states, but also for specific actions taken within those states, enabling the agent to evaluate the quality of particular actions in given situations. One of the biggest challenges in RL is designing an effective reward function~\citep{712192,ma2023eureka}. A poorly designed reward system can mislead the agent, resulting in suboptimal or unintended behaviors. Crafting rewards that properly guide the agent toward good outcomes is crucial. Other two significant challenges are representing the state space and exploring it. If states are not represented accurately or comprehensively, the agent may struggle to learn the true dynamics of the environment, which can impede its ability to make optimal decisions~\citep{mnih2013playingatarideepreinforcement}.

This section reviews key works that exemplify how generative models have been employed to address critical aspects of the RL training loop in robotics, specifically focusing on: (i) \textit{Reward Signal} generation in the presence of sparse rewards and under goal specification constraints, (ii) \textit{State Representation} learning to improve sample efficiency, and (iii) \textit{Planning \& Exploration} mechanisms that support generalization to new tasks. Based on the modality and purpose of the generative models employed in relation to the RL \textit{Task}, the works are categorized and examined. A more thorough classification of the basic models under examination is provided in Table~\ref{tab:tools}.

\begin{table}[!htbp]
\centering
\tiny
\begin{tabular}{lp{2.5cm}p{2.5cm}p{2.5cm}p{2.5cm}}
\toprule
\textbf{Network} & \textbf{Reward Signal} & \textbf{State Representation} & \textbf{Planning \& Exploration} \\
\midrule

\rowcolor{gray!10}
\textbf{LLM} & 
\makecell[l]{Textual\\ feedback} & 
\makecell[l]{Textual state\\ encoding} & 
\makecell[l]{High-level\\ guidance} \\

\rowcolor{white}
\textbf{VLM} & 
\makecell[l]{Multi-modal\\ feedback} & 
\makecell[l]{Multi-modal\\ state encoding} & 
\makecell[l]{Multi-modal\\ evaluation} \\

\rowcolor{gray!10}
\textbf{Diffusion Model} & 
\makecell[l]{Reward\\ generation} & 
\makecell[l]{State\\ generation} & 
\makecell[l]{Action\\ sampling} \\

\rowcolor{white}
\textbf{WM and VPM} & 
\makecell[l]{Reward\\ prediction} & 
\makecell[l]{State\\ prediction} & 
\makecell[l]{Model-based\\ planning} \\

\bottomrule
\end{tabular}
\vspace{0.2cm}
\captionsetup{justification=centering}
\caption{\textbf{Generative AI tools for RL integration.} Classification of generative AI models as tools based on integration into RL, as anticipated in Section~\ref{sec:basemodel} and Section~\ref{sec:modality}.}
\label{tab:tools}
\end{table}

\subsection{Reward Signal}
An increasing amount of research explores how generative AI models, particularly foundation models, can help automate or enhance reward design~\citep{liu2024rl,xietext2reward,song2023self,colas2023augmenting,du2023guiding,triantafyllidis2024intrinsic,carta2023grounding,Huang2023GroundedDG,ma2023eureka,chu2023acceleratingRL,bhateja2023robotic,kumar2022pre,adeniji2023language,cui2022can,lee2024affordance,mahmoudieh2022zero,rocamonde2023vision,baumli2023vision,chen2021learning,escontrela2024video}. These models offer new ways to interpret user input, perceive environments, and produce reward signals that guide effective policy learning.

\subsubsection{Reward design with LLMs}
Recently, LLMs such as GPT-4~\citep{achiam2023gpt} or Llama 3~\citep{dubey2024llama}, have been used to address the challenges of reward function design and direct reward specification. While LLMs struggle to directly control robots due to their ineffectiveness to produce control commands, they are highly valuable in evaluating the performance of RL agents, performing in-context reasoning on natural language problems. Traditionally, reward functions in RL are meticulously hand-crafted by researchers, often requiring significant domain expertise and limiting the agent's ability to adapt to new situations.  However, foundation models offer a compelling alternative by providing rich, generalist knowledge representations that can be leveraged to automatically generate or inform reward functions. Imagine a robot tasked with ``cleaning the kitchen''. An LLM could identify sub-tasks like ``wiping the table'' and ``putting dishes in the dishwasher'' and assign rewards based on their successful completion.  This structured approach breaks down complex tasks and provides clear goals for the RL agent.

Some approaches use LLMs for zero-shot reward generation~\citep{ma2023eureka,xietext2reward,song2023self} to convert natural language descriptions of desired behaviors into mathematical reward functions. TEXT2REWARD~\citep{xietext2reward} and Self-Refined LLM~\citep{song2023self} use LLMs to generate reward functions as executable code, while EUREKA~\citep{ma2023eureka} employs LLMs for evolutionary optimization of reward code. These methods leverage the LLM's ability to understand task semantics and translate them into reward functions. Others leverage the in-context learning capabilities of LLMs to learn from demonstrations or feedback provided in natural language, shaping the reward function iteratively~\citep{colas2023augmenting,du2023guiding,triantafyllidis2024intrinsic,chu2023acceleratingRL,yu2023language}. For example, Lafite-RL~\citep{chu2023acceleratingRL} uses LLM feedback to iteratively refine the reward function, employing few-shot learning to guide the LLM's generation process; while Language to Rewards~\citep{yu2023language} uses LLM feedback to define reward parameters that are optimized for specific tasks. While TEXT2REWARD and Lafite-RL focus on generating dense reward functions that provide feedback at each timestep, other methods, like Language to Rewards, may generate sparse rewards that are only provided at the end of an episode. Some methods, such as EUREKA and Language to Rewards, incorporate interactive feedback from humans to refine the reward function. In FoMo Rewards \citep{lubana2023fomo}, LLMs are employed to evaluate the likelihood of an instruction accurately describing a task given a trajectory of observations. This likelihood then serves as a reward signal. While the use of LLMs for reward function generation is still an emerging field, it shows significant promise for improving the development and deployment of intelligent agents. This approach can streamline the reward function design process, making it more general and task agnostic. Lastly, \citet{nair2022learning} showcase the use of LLMs to learn language-conditioned rewards from offline data and crowd-sourced annotations. This approach avoids the need for extensive human demonstrations, offering a scalable solution for learning complex behaviors. These different approaches highlight the versatility of LLMs in reward generation for RL in robotics, while the choice of approach depends on the specific requirements of the task and the available resources. However, one main drawback we found in existing work leveraging LLMs to craft rewards is that, since they work with text-only input, they are limited in acquiring perception from the real world~\citep{ma2024dreureka}. This especially limits their capability to simulation environments where full state information is readily available.

\subsubsection{Reward design with VLMs}
The integration of vision-language models~\citep{radford2021learning,ramesh2021zero} in RL task evaluation represents a significant advancement over methods that rely solely on textual or numerical feedback~\citep{mahmoudieh2022zero,rocamonde2023vision,adeniji2023language,baumli2023vision,sontakke2024roboclip,wang2024,adeniji2023language,cui2022can,lee2024affordance,yang2024robot}. By leveraging the power of visual understanding, VLMs can be used for task evaluation \citep{adeniji2023language,cui2022can} based on prompts and image frames as input.

Many robotic tasks involve interacting with a complex environment filled with diverse sensory information. Vision foundation models can leverage their ability to process different data modalities (text, images) to construct richer reward functions. Imagine a robot tasked with sorting laundry. A VLM could analyze an image of the clothing to identify the fabric type and color, and combine this information with text instructions to create a reward function that promotes sorting based on pre-defined categories, performing multi-modal reward learning.
In \citet{mahmoudieh2022zero}, \citet{rocamonde2023vision} and \citet{adeniji2023language} VLMs are employed as zero-shot reward models. These approaches leverage the in-context learning ability of VLMs to evaluate if a visual scene coming from a camera sensor effectively match the language task description. This approach eliminates the need for manually designing reward functions or gathering extensive human feedback; moreover, it does not need environment state information. Differently, in RoboCLIP~\citep{sontakke2024roboclip} rewards are generated by evaluating how closely the robot's actions match the provided example. Interestingly, Code as Reward \citep{venuto2024code} utilizes VLMs to generate reward functions through code, thereby reducing the computational overhead of directly querying the VLM every time. \citet{wang2024} takes a slightly different route, querying VLMs to express preferences over pairs of image observations based on task descriptions, and subsequently learning a reward function from these preferences.

\subsubsection{Reward design with diffusion models}
Similar to other types of generative AI tools, diffusion models can be leveraged to infer rewards in RL. \citet{Huang2023DiffusionReward} and \citet{nuti2024extracting} present methods for deriving reward functions by conditional diffusion models that are trained on expert visual demonstrations to generate rewards, similarly to the usage of diffusion models for conditional image generation~\citep{rombach2022high}. On a different approach, \citet{mazoure2023value} introduces the Diffused Value Function (DVF) algorithm, which leverages diffusion models to estimate value functions from states and actions, enhancing efficiency without explicitly learning rewards. They show improved long-term decisions. To conclude, diffusion models are a relatively new approach to reward design, but they show strong promise; particularly due to adaptability, dense reward generation, and fast inference. We do not yet see immediate applications for text-conditioned reward generation.

\subsubsection{Reward design with video prediction}
The approaches for reward design presented so far in our taxonomy often lack accuracy in predicting changes in the robot's visual state within the environment. Recent methods, such as those in \citet{chen2021learning} and \citet{escontrela2024video}, utilize the ability to predict future video frames based on current observations to shape rewards more effectively, leveraging environmental models (\textit{e.g.}, video prediction models) to enhance reward specification in RL systems. Specifically, \citet{chen2021learning} propose DVD, a discriminator that learns multi-task reward functions by classifying whether two videos perform the same task. This approach allows generalization to unseen environments and tasks by learning from a small amount of robot data and a large dataset of human videos. \citet{escontrela2024video} introduce VIPER, which uses pre-trained VPMs to provide reward signals for RL based on frame observations rather than the actions the robot takes. VIPER achieves expert-level robot control across various tasks in simulations without predefined rewards, demonstrating the potential of VPMs for online reward specification during robotic task execution. At the same time, other researchers propose HOLD~\citep{alakuijala2023learning} and VIP~\citep{ma2022vip}. This approaches generalize to unseen robot embodiments and environments, effectively accelerating RL training on various manipulation tasks zero-shot; the second one is trained on large-scale human videos.

\subsection{State Representation}
Foundation models can offer a promising avenue to address the challenge of more realistic state representations for data augmentation and training in RL~\citep{chen2023open,nottingham2023embodied,liu2024rl,nottingham2023embodied,Wang2023GenSimGR,yang2023learning,seo2023multi,wang2023robogen,mazzaglia2024multimodal,wu2024ivideogpt,bruce2024genie}. However, the successful integration of foundation models as world representations for meaningfully training robots on augmented data and through model-based RL techniques hinges on addressing a fundamental issue mentioned many times in our discussion: \textit{grounding}. While these models excel at processing abstract representations, their ability to effectively model realistic scenarios depends on establishing a meaningful connection between these representations and the real-world objects or concepts they abstract~\citep{wu2023daydreamer}. This section surveys research papers that utilize learned representations to predict and interact with environments in RL.

\subsubsection{Learning representations from videos}
Integrating VPMs with RL enhances the training of robots by allowing them to predict future frames from a sequence of previous images. VPMs forecast how a state will evolve, enabling the RL agent to plan actions and make decisions with more data-efficient learning and safer exploration, by anticipating future scenarios while reducing reliance on real-world trials~\citep{du2023learning,ye2023foundation,bhateja2023robotic,majumdar2024we}. However, the success of this approach depends on the accuracy of the predictions, as well as managing the increased computational complexity, that often limits the possibility to use a single pre-trained model for a wide variety of robotics tasks. As a result, a common trend in research is to train or fine-tune different VPMs for specific applications, but future work could focus on scaling VPMs to improve their generalization.

\citet{du2023learning} present UniPi, a unique approach that casts sequential decision-making as a text-conditioned video generation problem. UniPi leverages the knowledge embedded in language and videos to generalize to novel goals and tasks across diverse environments and to transfer effectively to downstream RL tasks. This approach enables multi-task learning and action planning, showcasing the potential of video generation for policy learning. Foundation RL \citep{ye2023foundation} takes this objective a step further, they trained an RL model that leverages foundation priors from large-scale pre-training for embodied agents and they tested it on simulated robotic tasks.

\subsubsection{Foundation world models for model-based RL} 
Many RL algorithms rely on accurate models of the environment's dynamics to plan and make decisions~\citep{712192}. World models \citep{ha2018recurrent}, particularly those trained on data that includes physical interactions, can be used to learn these models of the world. The learned model can help the RL agent to predict the consequences of its actions (\textit{e.g.}, its future states) and make informed decisions based on predictions~\citep{wu2023daydreamer}.

World models capture the action space of a robot. They can be constructed in a variety of ways, utilizing diverse data sources and employing different learning architectures. One recent approach to building a world model for abstract representation is through the use of large language models~\citep{nottingham2023embodied,Zala2024EnvGen,Wang2023GenSimGR,seo2023masked,seo2023multi}. For example, \citet{Zala2024EnvGen} propose EnvGen, a framework where an LLM generates and adapts training environments for small RL agents. By creating diverse environments tailored to specific skills, EnvGen enables agents to learn more efficiently in parallel. The LLM receives feedback on agent performance and iteratively refines the environments, focusing on weaker skills. On a similar approach \citet{Wang2023GenSimGR} introduces GenSim, which uses GPT-4~\citep{achiam2023gpt} to automate the generation of diverse simulation tasks. Differently, \citet{yang2023learning} further advance the state-of-the-art in learning interactive real-world simulators for robotics as foundation models. They leverage diverse datasets, each rich in different aspects of real-world experience, to simulate the visual outcomes of both high-level instructions and low-level controls. The resulting simulator is used to train policies that can be deployed in real-world scenarios, showcasing the potential of bridging the sim-to-real gap in embodied learning.

World models can also be trained at a very large scale from scratch, leading to the development of multi-modal foundation world models capable of generalizing greatly in vertical domains. Building on this for efficient robot training, \citet{wang2023robogen} present RoboGen, a generative robotic agent that automatically learns diverse skills through environment simulation. Similarly, \citet{mazzaglia2024multimodal} introduce GenRL, a framework for model-based RL training of generalist agents, that learns a multi-modal foundation world model. Moreover, \citet{wu2024ivideogpt} presents iVideoGPT, a scalable autoregressive transformer framework for interactive world models. It is pre-trained on millions of human and robotic manipulation trajectories, demonstrating its versatility in various downstream tasks. While \citet{mao2024zero} focuses on zero-shot safety prediction for autonomous robots using foundation world models. They propose a world model that combines foundation models with interpretable embeddings, addressing the distribution shift issue in standard world models. Their approach demonstrates superior state prediction and excels in safety predictions, highlighting the potential of foundation models for safety-critical applications. Lastly, \citet{bruce2024genie} introduces Genie, the first generative interactive environment trained in an unsupervised manner from unlabeled Internet videos.

While significant progress has been made in model-based RL and world models could unlock unlimited data availability for training, \citet{wolczyk2023role} discuss the issue of catastrophic forgetting in post-training RL models, where pre-trained knowledge can be lost as new tasks are learned. This problem is particularly evident in compositional tasks, where different parts of the environment are introduced at different stages of training.

\subsection{Planning \& Exploration}
Policy learning refers to the process of determining a set of actions that can be executed to reach the desired target state for a given task~\citep{sutton1988learning,gu2022review}. In this section, we briefly highlight interesting findings in the literature where generative AI models are used to learn effective policies for RL. In particular, two important concepts are exploration, which enables the agent to discover new states and maximize rewards, and planning, which involves combining skills to develop more comprehensive policies~\citep{liu2024rl,mezghani2023think,psenka2023learning}. Various approaches leveraging LLMs, VLMs, and diffusion models have been explored to enhance both exploration and planning in RL settings~\citep{colas2020language,hu2023instructed,zhang2023bootstrap,chen2024vision,ma2024explorllm,cui2022can,liang2023adaptdiffuser,zhou2024adaptive,brehmer2024edgi}.

\subsubsection{LLMs for planning and exploration}
LLMs can be used for planning in RL, where the LLM acts as a strategic semantic planner, guiding the application of learned RL skills to new tasks based on task-specific prompts. \citet{ahn2022can} and \citet{Huang2023GroundedDG} introduce two methods that combines LLMs with pre-trained skills and affordance functions extracted from the RL training to ground language in robotic actions. The LLM is used to propose high-level actions, while the affordance functions, often learned through RL, determine the feasibility of these actions in the current context. This approach enables robots to execute complex tasks based on natural language instructions by ensuring that the proposed actions are both semantically relevant and physically feasible. Similarly, Plan-Seq-Learn (PSL)~\citep{dalal2024plan}, is a modular approach that uses motion planning to connect abstract language from LLMs with learned low-level control for solving long-horizon robotics tasks. PSL breaks down tasks into sub-sequences, uses vision and motion planning to translate these sub-sequences into actionable steps, and then employs RL to learn the necessary low-level control strategies. This approach enables robots to efficiently learn and execute complex tasks by leveraging the strengths of both LLMs and RL.

Regarding exploration, \citet{colas2020language} were the first to work on it by specifically incorporating language-driven imagination into RL. According to their IMAGINE architecture, ``imagination'' is the process of applying a learned goal recognizer to relabel prior experiences with other, language-based objectives. By predicting which natural language descriptions could realistically correspond to an episode's outcome, this methodology enables the agent to associate new hypothetical objectives with previous ones. The agent may efficiently learn from these relabeled goals by combining a modular policy with a language-conditioned reward function. This improves generalization and exploration in a variety of language-specific tasks. The research underscores the critical role of language in enhancing creative, goal-driven exploration in RL. LLMs have shown promise in directly generating RL policies, instead of generating reward functions: GLAM \citep{carta2023grounding}, InstructRL \citep{hu2023instructed}, and BOSS \citep{zhang2023bootstrap}, explore this, each with distinct approaches and contributions. GLAM focuses on grounding LLMs in interactive environments through online RL. It utilizes an LLM as the policy for an agent operating in a textual environment, refining the LLM's understanding of the environment through continuous interaction and feedback. This approach aims to address the challenge of aligning the LLM's knowledge with the actual environment dynamics, improving its ability to make decisions and achieve goals. The key innovation of GLAM lies in its online learning approach, where the LLM is not just pre-trained on existing data but actively learns and adapts as it interacts with the environment. This allows for a more dynamic and context-aware policy generation process. InstructRL, on the other hand, introduces a framework where humans provide high-level natural language instructions to guide the agent's behavior. These instructions are used to generate a prior policy using LLMs, which then regularizes the RL objective. This approach aims to align the agent's actions with human preferences and expectations, making it more suitable for collaborative tasks. InstructRL bridges the gap between human intentions and agent actions. The authors acknowledge that their work is limited by the challenges of abstracting certain actions, like continuous robot joint angles, into language, but they are optimistic about future advancements in multi-modal models expanding their applicability. Addressing the problem of generating realistic robotic policies and providing a more robust framework for translating abstract concepts into precise commands for robots, BOSS tackles the challenge of learning long-horizon tasks with minimal supervision. It starts with a set of primitive skills and progressively expands its skill repertoire through a bootstrapping phase. During this phase, the agent practices chaining skills together, guided by LLMs that suggest meaningful combinations. This approach enables the agent to learn complex behaviors autonomously, reducing the need for extensive human demonstrations or reward engineering. BOSS's innovation lies in its ability to leverage the knowledge embedded in LLMs to guide the exploration and learning of new skills, making it a promising approach for developing generalist agents capable of performing a wide range of tasks.

\subsubsection{VLMs for exploration}
Another major theme in the research is the use of FMs that work with images together with text to guide exploration in RL agents. VLMs offer deeper grounding in robotics applications compared to LLMs because they directly associate visual perception with the input task in text form and do not require a separate vision module for perception, hence enhancing efficient exploration based on goal evaluation~\citep{ma2024explorllm,ma2023liv,cui2022can}. By processing real-time sensory data from the robot (\textit{e.g.}, camera images), the model can identify unexpected situations or deviations from the expected plan. This information can be used to modify the reward function online, penalizing actions that lead to undesirable outcomes and encouraging exploration of alternative strategies. For example, if a robot attempting to pick up a cup encounters an obstacle, the foundation model could adjust the reward function to prioritize navigating around the obstacle before resuming the grasping attempt. As demonstrated by \citet{cui2022can}, VLMs can enable zero-shot task specification in robotic manipulation by supporting more general and user-friendly goal representations---such as internet images or hand-drawn sketches---which promote exploration more closely aligned with the intended task. Other studies, however, assert that general-purpose VLMs might struggle with exploration and result in agents that act too roughly, especially in online RL~\citep{chen2024vision}.

\subsubsection{Diffusion models for planning and exploration}
Conventional RL planning techniques frequently use deterministic algorithms, which might perform poorly in complicated or unpredictable contexts~\citep{janson2018deterministic,duchovn2014path,zhou2022review}. Diffusion models, which were first created for generative tasks, have been modified to provide flexibility and stochasticity to the planning process in order to add robustness~\citep{liang2023adaptdiffuser,zhou2024adaptive,brehmer2024edgi,li2023hierarchical,he2024diffusion}.
These models allow for flexible decision-making by iteratively converting noise into planned action sequences. One important strategy, Diffuser \citep{janner2022diffuser}, combines RL with diffusion models to create reward-conditioned plans based on prior knowledge. It is excellent at long-term planning. To improve sample efficiency and generalization, extensions such as EDGI~\citep{brehmer2024edgi} treat planning as conditional sampling and add domain constraints. Other studies focus on safe planning, guaranteeing constraint satisfaction, and refining plans that are not feasible~\citep{xiao2023safediffuser,lee2024refining}.
Diffusion models also improve exploration by stochastically generating a wide range of possible states and objectives~\citep{wang2023cold,li2023hierarchical,zhang2024language,kim2024stitching,chen2024simple}.

Lastly, diffusion models for planning and exploration works well in \textbf{offline RL}, where we can train diffusion models to generate high-quality actions from large datasets or benchmarks~\citep{fu2020d4rl,lu2023contrastive,suh2023fighting,kang2024efficient,hu2023instructed,hansen2023idql,psenka2023learning,jain2023learning,zhu2023madiff,kim2024robust}. Some methods~\citep{suh2023fighting,lu2023contrastive}, use gradient-based planning or energy functions to steer action generation and improve reward outcomes. Other works~\citep{kang2024efficient}, focus on optimizing the sampling process to make diffusion models more practical and faster. Researchers also developed techniques like Implicit Diffusion Q-Learning~\citep{hansen2023idql} and Q-Score Matching~\citep{psenka2023learning}, incorporating Q-learning ideas to better connect action choices with predicted rewards training offline~\citep{hu2023instructed,ho2022classifier,venkatraman2023reasoning,rombach2022high,jain2023learning}.

\section{RL Pre-Training}
\label{sec:genpolicies}
Until now, our survey has examined Generative AI as a tool within the RL pipeline (\textit{Generative Tools for RL}). We now turn the coin over and explore the converse perspective: using RL itself to \textit{pre-train}, \textit{fine-tune}, and \textit{distill} generative policy models. In \textit{RL Pre-Training} we survey methods that pre‑train transformer‑ or diffusion‑based policies with RL objectives.

\subsection{Transformer Policy}
Autoregressive transformer models were originally restricted to the domain of natural language, but they now form an integral part of RL, especially when the task can be framed as a sequential decision-making problem~\citep{hu2024harmodt,bucker2023latte,li2024online,wolczyk2024fine,reed2022generalist}. In particular, specialized adaptations of the Transformer architecture, such as Decision Transformers (DTs)~\citep{chen2021decision}, are becoming more popular. Like language models, DTs generate tokens (\textit{i.e.}, actions) one at a time, conditioned on the previous context, and use the Transformer to predict the next action in a sequence based on previous return, state, and action tuples.
Recent transformer-based RL studies focus on scalability, adaptivity, and transferability~\citep{hu2023prompt,kumar2022pre,wen2022multi,xu2023hyper}. Results from \citet{wen2022multi} and \citet{xu2023hyper} show how large models may generalize across various RL tasks. These methods show how transformer backbones can learn from little supervision and generalize well.
HarmoDT~\citep{hu2024harmodt}, LATTE~\citep{bucker2023latte}, and PACT~\citep{bonatti2023pact} would be examples of such architectures that have trained transformers at scale for RL-based robotic control on real robots. Tackling a different problem, Q-Transformer~\citep{chebotar2023q} leverages offline data, human demonstrations, and trajectory rollouts to learn Q-functions with transformers so that robots can perform tasks appropriately. AnyMorph~\citep{trabucco2022anymorph} pushes the generalization frontier by adapting to different robot morphologies. We describe here the small number of DT variants applicable to robotics, but we envision that many RL problems may be addressed by foundation models utilizing DTs backbones~\citep{wen2023large}. Lastly, a major trend is merging language and reasoning into policies: \citet{mezghani2023think} merge language generation with action prediction, allowing agents to reason in natural language while planning actions. This has not been validated directly in real-world robotics but gives an interesting approach for tasks which require long-term planning. Recently, GATO~\citep{reed2022generalist} emerged as a generalist agent due to its ability to execute tasks across modalities, spanning from gaming to real-world robotics, using a single transformer backbone that performs input fusion into a RL policy.

\begin{table}[!htbp]
\centering
\tiny
\begin{tabular}{lp{4.5cm}p{4cm}}
\toprule
\textbf{Aspect} & \textbf{Transformer Autoregression} & \textbf{Diffusion Model} \\
\midrule

\rowcolor{gray!10}
\textbf{Generation Process} & 
Sequential, one step at a time. & 
Iterative, gradual refinement from noise. \\

\textbf{Output Type} & 
Typically discrete (e.g., tokens, discrete actions). & 
Typically continuous (e.g., trajectories, continuous actions). \\

\rowcolor{gray!10}
\textbf{Training Objective} & 
Maximize likelihood of observed data. & 
Predict noise added during forward diffusion. \\

\textbf{Inference Speed} & 
Faster (if same size). & 
Slower due to iterative refinement. \\

\rowcolor{gray!10}
\textbf{Error Propagation} & 
Errors can compound over time. & 
Less prone to compounding errors. \\

\textbf{Expressiveness} & 
Highly expressive for sequential data. & 
Highly expressive for continuous data. \\

\rowcolor{gray!10}
\textbf{Applications} & 
Discrete action RL policies. & 
Continuous action RL policies. \\

\textbf{RL Fine-tuning Ease} & 
Straightforward with standard RL algorithms (e.g., PPO, Q-learning). & 
More complex, may require custom integration with RL. \\

\rowcolor{gray!10}
\textbf{Sample Efficiency} & 
Moderate; improves with pretraining. & 
High; performs well in low-data regimes. \\

\textbf{Action Modeling} & 
Limited multi-modality; can struggle in complex spaces. & 
Strong multi-modal capabilities; excels in complex control. \\

\rowcolor{gray!10}
\textbf{Best For} & 
Long-horizon, complex reasoning tasks; fast inference scenarios. & 
Smooth, continuous control; planning; multi-modal outputs. \\

\bottomrule
\end{tabular}
\vspace{0.2cm}
\captionsetup{justification=centering}
\caption{\textbf{Comparison of generative policies.} Transformer autoregressive policies versus diffusion non-autoregressive policies in RL.}
\label{tab:transformer_vs_diffusion}
\end{table}

\subsection{Diffusion Policy}
Using iterative refinement instead of step-by-step prediction, diffusion models have recently acquired popularity as alternative to autoregressive techniques in robotic policy generation~\citep{chi2023diffusion,janner2022diffuser} (see Table~\ref{tab:transformer_vs_diffusion} for a comparison). They are ideal for dynamic robotic environments due to their ability to manage uncertainty and provide a variety of behaviors~\citep{huang2024diffuseloco,zhu2023diffusion,wang2022diffusion}. Diffusion models can model complicated or high-dimensional dynamics by refining probability distributions to construct action sequences, in contrast to classic RL methods that yield deterministic outputs~\citep{712192,he2024diffusion}. They can incorporate constraints such as safety or energy efficiency and allow expressive, context-sensitive behaviors by directly modeling action distributions~\citep{ni2023metadiffuser,li2023beyond}.
\citet{hegde2023generating} present a method to condense a large archive of policies, trained using RL, into a single generative model. A diffusion model is then trained on these compressed representations to generate new policies conditioned on specific behaviors, either through quantitative measures or language descriptions.
Other studies use diffusion models as policy in offline RL, improving performance and scalability~\citep{chen2023score,ding2023consistency}. Moreover, Diffusion-QL~\citep{wang2022diffusion} uses conditional diffusion to match IL with Q-learning while preserving demonstration data proximity and optimizing rewards.
Lastly, frameworks such as Decision Diffuser~\citep{ajay2022conditional} and DIPO~\citep{yang2023policy}, formulate the sequential decision-making problem as a conditional generative modeling problem using RL to train a diffuser.

\section{RL Fine-Tuning}
\label{sec:finetuning}
Generative models, such as transformer-based and diffusion-based policies, are often trained with IL and they demand flexible RL fine-tuning strategies, if expert data for supervised fine-tuning (SFT) are limited. Recent works effectively integrate these expressive models into RL pipelines without instability or loss of efficiency~\citep{FLaReRL2024,mark2024policy}.
In this section, we explore two major classes of generative policies where we identified RL fine-tuning is particularly relevant, (i) large transformer policies (mostly VLA models), and (ii) diffusion policies (usually smaller in size).

\begin{figure}[!htbp]
    \centering
    \begin{subfigure}[t]{0.48\textwidth}
        \centering
        \includegraphics[width=\textwidth]{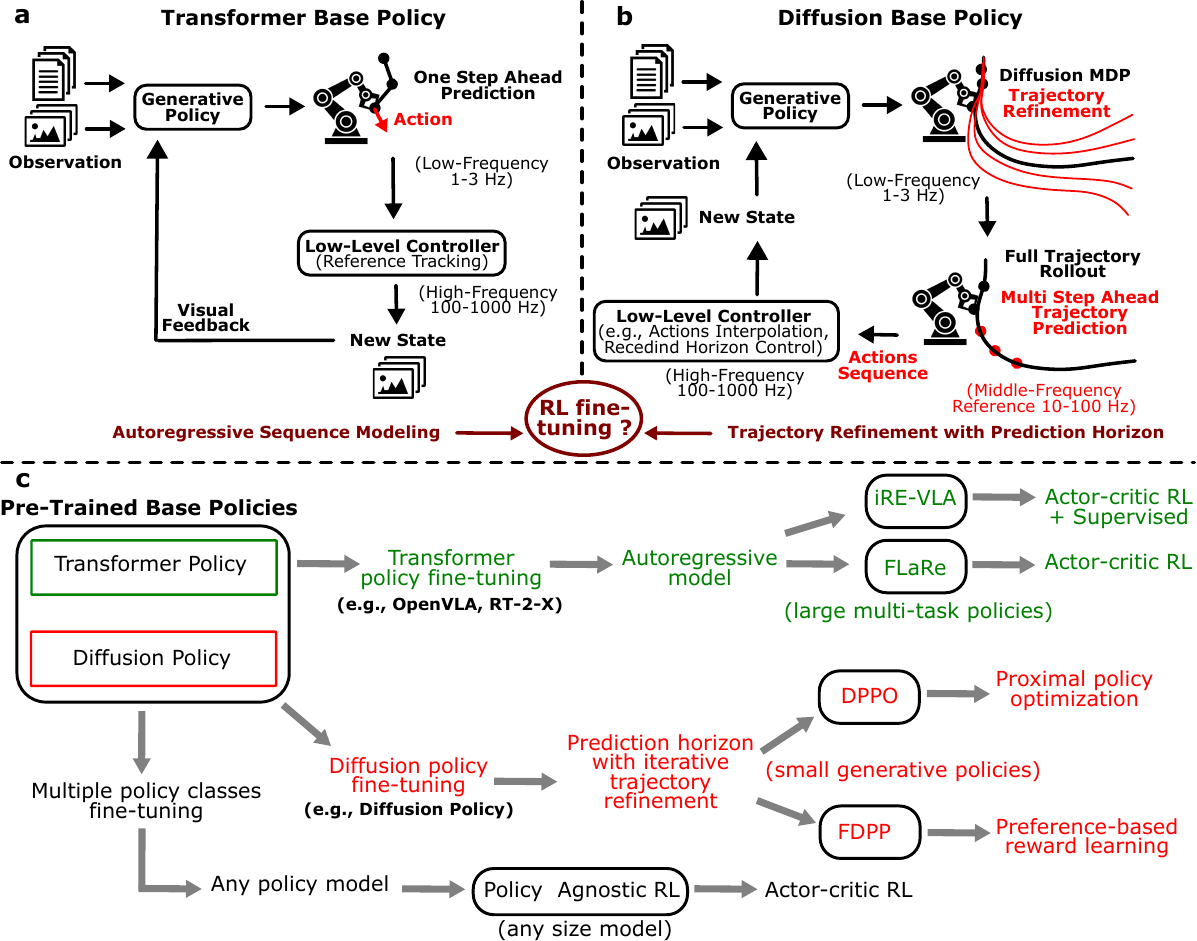}
        \captionsetup{justification=centering}
        \caption{This figure illustrates the transformer policy (panel~\textbf{a} shows VLA inference time) and the diffusion policy (panel~\textbf{b}) principles. In panel~\textbf{c} we show recent RL methods that can be used to fine-tune a base generative policy.}
        \label{fig:genpolicies}
    \end{subfigure}
    \hfill
    \begin{subfigure}[t]{0.48\textwidth}
        \centering
        \includegraphics[width=\textwidth]{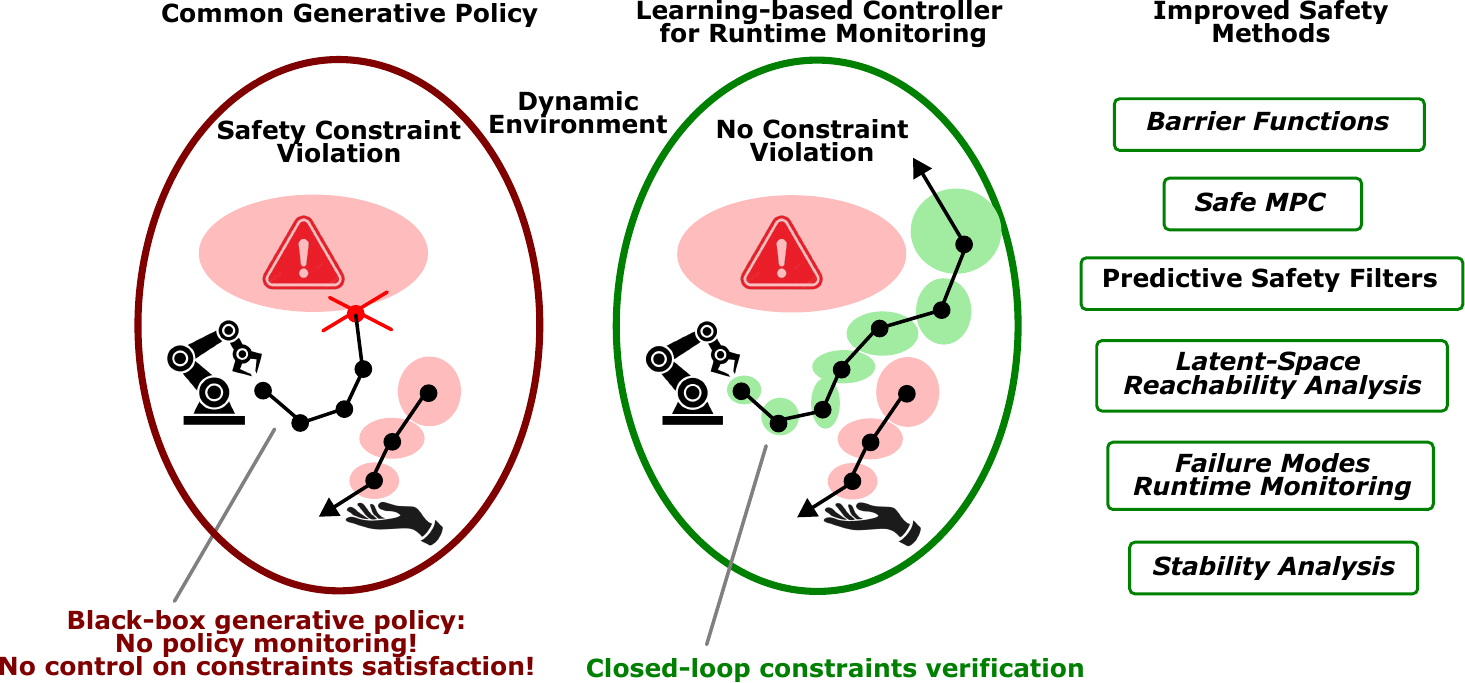}
        \captionsetup{justification=centering}
        \caption{This figure illustrates the challenge of deploying a black-box generative policy in real-world dynamic environments without access to model-based information.}
        \label{fig:safety}
    \end{subfigure}
    \captionsetup{justification=centering}
    \caption{\textbf{Generative policies.}}
    \label{fig:genpoliciesandsafety}
\end{figure}

Recent advances in \textbf{Transformer-based policies} have transformed robotic control, with robotics foundation models excelling through large-scale pre-training. A major step toward generative policy architectures and datasets for robotics is the OpenX Embodiment project \citep{open_x_embodiment_rt_x_2023}, which introduced the largest open-source real-robot dataset---the Open X-Embodiment Dataset. This dataset supports the development of pivotal VLA models like RT-X models~\citep{brohan2022rt,brohan2023rt}, Octo~\citep{octo_2023}, and OpenVLA \citep{kim2024openvlaopensourcevisionlanguageactionmodel}. However, state-of-the-art generalist VLA policies are typically trained via IL and often require additional training to adapt to out-of-distribution scenarios~\citep{li24simpler,ma2024surveyvisionlanguageactionmodelsembodied,xiao2023robotlearningerafoundation,hu2023generalpurposerobotsfoundationmodels}. RL has become an alternative to improve policy performance, particularly when it comes to managing distribution shifts or leading generalist models toward specialization. This has motivated the development of new RL-based fine-tuning algorithms, particularly actor-critic methods~\citep{guo2025improving,liu2025can}. Actor-Critic approaches enable very reliable and effective policy updates by utilizing two networks: the critic to assess actions and the actor to decide on them. This method aids in RL fine-tuning by continuously enhancing action selections in response to critic feedback~\citep{712192}. Through the use of on-policy algorithms such as PPO, cautious learning rates, and distinct actor-critic networks, the FLaRe framework~\citep{FLaReRL2024} demonstrates RL-based fine-tuning by matching pre-trained transformer policies with new experience. FLaRe reduces the sim-to-real gap by using domain randomization and extensive simulations, increasing success rates by roughly +30 \% on both real robots and in simulation. PPO is the recommended RL technique for VLA fine-tuning, according to \citet{liu2025can}. Its advantages include shared actor-critic backbones, short training epochs, and great generalization on unseen objects and dynamic situations. However, the low-frequency action generation (reported in panel \textbf{a} in Figure~\ref{fig:genpolicies}) of VLA models and the requirement of large amounts of interaction data even for minor adjustments, make online fine-tuning difficult and limited in rel world settings. Standard fine-tuning methods still need further exploration in case of VLA models, and current research focuses on large-scale RL fine-tuning for transformer policies~\citep{FLaReRL2024}.

An alternative area where RL fine-tuning shows potential is with \textbf{Diffusion-based policies}, which offer many possibilities with RL due to their ability to predict smooth trajectories non-autoregressively at each inference step~\citep{wang2022diffusion,chen2025fdpp}. However, key problems in applying RL to diffusion policies are include the high-dimensional nature of the action space, the slow iterative denoising process involved in action generation, and training instability. Standard policy gradient methods have been viewed as inefficient for such models, due to the increased number of steps over which rewards must be predicted, introduced by iterative denoising~\citep{yang2023policy}. Recent advances, such as Diffusion Policy Policy Optimization (DPPO), have demonstrated that RL can be effectively integrated with diffusion policies by formulating the denoising process as a Markov Decision Process~\citep{dppo2024}. This approach enables policy gradients to propagate through the diffusion steps, leveraging the structured noise removal process to facilitate more stable training.
A major advantage of RL-fine-tuned diffusion policies is their ability to engage in on-manifold exploration, meaning that the policy remains close to the expert data distribution while still improving performance through RL. This structured exploration contrasts with traditional RL methods, which often struggle with off-manifold exploration, leading to unstable training and suboptimal policies~\citep{DiffOptPermenter2024}. Panel~\textbf{b} of Figure~\ref{fig:genpolicies} shows the working principle of diffusion policies, while panel~\textbf{c} depicts the state-of-the-art methods for RL-based fine-tuning generative policies.

In general, most research focuses on transformer-based or diffusion-based policy architectures separately, as they align with different goals and fine-tuning methods, although Actor-Critic RL is commonly used. While some researchers are exploring these pathways, others aim to develop a generalized RL-based fine-tuning framework for any generative policy, regardless of size and backbone model~\citep{mark2024policy}.

\section{Policy Distillation}
\label{sec:distillation}
Policy distillation is a well-known concept in the RL literature~\citep{rusu2015policy}. Recently, with the emergence of large generalist generative policies, RL-based methods for policy distillation have been applied to VLA models. In particular, recent works have explored policy distillation in OpenVLA~\citep{kim2024openvlaopensourcevisionlanguageactionmodel} and Octo~\citep{octo_2023}.
The goal of policy distillation is to transfer pre-trained knowledge from a \textit{teacher policy} to a \textit{student policy}. The recent literature can be categorized into two opposing approaches:
{\small
\begin{itemize}[noitemsep, topsep=0pt, leftmargin=*]
    \item \textbf{From Generalist to Expert:} \citet{julg2025refined} developed a method to create task-specific RL agents distilled from a pre-trained VLA model. While VLA models are known for their strong generalization capabilities, they often struggle to achieve highly precise results compared to task-specific RL policies. The key idea is that the internal knowledge of VLA models can be useful in guiding RL exploration for specific agents. However, this work primarily presents preliminary simulation results, emphasizing that sim-to-real transfer remains the main challenge when training RL policies in simulation. This is particularly relevant given that OpenVLA and Octo typically perform better in real-world conditions \citep{open_x_embodiment_rt_x_2023,li24simpler}.
    \item \textbf{From Experts to Generalist:} \citet{xu2024rldg} propose a method to improve OpenVLA and Octo using demonstrations from expert RL policies. Their approach focuses on generating high-quality training data to fine-tune VLA models, as these models performance is highly dependent on the quality of their training data. Their experiments demonstrate that this method outperforms models trained solely on human demonstrations, achieving superior results in real-world precision manipulation tasks.
\end{itemize}
}
\section{Challenges, Recommendations and Future Research Directions}
\label{sec:challenges}
We organize challenges, recommendations and few important insights for the future into three macro-categories based on our taxonomy---Section~\ref{sec:c1} is related to challenges in \textit{Generative Tools for RL}, Section~\ref{sec:c2} presents challenges in \textit{RL for Generative Policies} and Section~\ref{sec:f1} is on \textit{Future Research Directions}. Each category addresses specific limitations and opportunities that must be tackled to enable scalable, safe, and adaptive robot learning that integrates generative AI and RL.

\subsection{Challenges in \textit{Generative Tools for RL}}
\label{sec:c1}
Every generative model we discussed has unique advantages for RL; consequently, different generative AI models present distinct challenges, and we highlight those we identified as most urgent to be addressed.

\subsubsection*{LLMs for reward signals}
Despite their strength in symbolic reasoning, LLMs lack grounding in real-world tasks~\citep{cohen2024survey,bhat2024grounding}. Main challenges are: (i) \textbf{limited contextual understanding}~\citep{ahn2022can,Huang2023GroundedDG,du2023guiding,colas2023augmenting,triantafyllidis2024intrinsic,dalal2024plan}, (ii) \textbf{real-world reduced adaptability}~\citep{ma2023eureka,ma2024dreureka,chu2023acceleratingRL,lubana2023fomo,xietext2reward} because most works are based on the assumption that the LLM has access to the environmental states from the simulation, and (iii) \textbf{symbolic-physical misalignment}~\citep{bhat2024grounding,cohen2024survey,bhat2024grounding,carta2023grounding}.

\textit{Recommendation:} Similarly to \citet{ma2024dreureka}, we believe that LLMs can improve reward generation in real-world. Develop architectures that pair LLMs with RL controllers capable of translating high-level goals into more controlled rewards. Employ LLM fine-tuning gradually increasing real-world sensor data exposure to avoid simulation environments dependency.

\subsubsection*{VLMs for reward signals and state representation}
VLMs excel at perception and language integration and they are more effective in real-world tasks than LLMs, but they struggle with translating visual-semantic representations into actionable RL policies. Challenges are: (i) \textbf{detailed scene understanding}~\citep{cui2022can,venuto2024code,ma2023liv,wang2024,yang2024robot,chen2024vision,mahmoudieh2022zero,ma2024explorllm} and (ii) \textbf{inferring physical affordances or dynamics}~\citep{huang2024dark,adeniji2023language,sontakke2024roboclip,di2023towards,rocamonde2023vision,baumli2023vision}.

\textit{Recommendation:} Associate VLMs with with more meaningful information to improve precision: extracting pose of relevant objects, keypoints, or region of interest in the frames. Use positive and negative visual examples to generate semantic rewards, since it's proved to be more effective than only positive ones by \citet{huang2024dark}. Use auxiliary tasks like affordance prediction to narrow the abstraction gap, and combine VLMs outputs with filtering mechanisms, as in \citet{ahn2022can}.

\subsubsection*{Diffusion models for planning and exploration}
Diffusion models offer strong distribution modeling and short-term planning, but suffer from: (i) \textbf{in-context understanding}~\citep{zhu2023diffusion,hansen2023idql,psenka2023learning,jain2023learning,chen2022offline,liang2023adaptdiffuser,zhou2024adaptive,wang2023cold,brehmer2024edgi}, (ii) \textbf{multi-modality extension}~\citep{suh2023fighting,kim2024robust,venkatraman2023reasoning,ni2023metadiffuser,he2024diffusion,janner2022diffuser}, and (iii) \textbf{poor reasoning}~\citep{lu2023contrastive,kang2024efficient,hu2023instructed,zhu2023madiff,li2023hierarchical,xiao2023safediffuser,chen2024simple}.

\textit{Recommendation:} Embed hierarchical control into diffusion models to manage long-horizon planning; integrate LLMs for reasoning levels with low-level diffusion-based modules. Combine diffusion models with RL through receding-horizon approaches for stable planning, as shown in recent approaches~\citep{zhu2023diffusion}. Guide diffusion models generation by conditioning them on state information or meaningful variables~\citep{ajay2022conditional}.

\subsubsection*{World models and video prediction for environment modeling}
World models and video prediction models give agents the ability to ``imagine'' future without actually interacting with the environment, which is particularly helpful in situations with limited data. However, they have drawbacks: (i) \textbf{restricted generalization}~\citep{chan2024offlinetoonline,ha2018recurrent,wu2024ivideogpt,mao2024zero,yang2023learning}, (ii) \textbf{compounding errors over extended rollouts}~\citep{van2024continual,seo2023masked,seo2023multi,nottingham2023embodied,Zala2024EnvGen}, and (iii) {misalignment between actionable state representations and visual accuracy}~\citep{Wang2023GenSimGR,mazzaglia2024multimodal,bruce2024genie,ye2023foundation,wang2023robogen}. Similar to this, video prediction algorithms try to predict changes at the pixel level, but they frequently produce (i) \textbf{misaligned forecasts}~\citep{chen2021learning} and have (ii) \textbf{poor physically significant dynamics}~\citep{escontrela2024video} over time.

\textit{Recommendation:} Use foundation world models~\citep{mazzaglia2024multimodal} rather than video predictors for more robust representation learning. Employ stochasticity and uncertainty modeling to mitigate compounding errors in prediction. Augment world models with high-fidelity sensory data~\citep{bruce2024genie,Wang2023GenSimGR}. Hybrid frameworks combining analytical simulators with learned models (\textit{e.g.}, \citep{Genesis}) can improve realism and transferability. Fine-tune video prediction models on task-specific environments and prefer them only for data augmentation or internal reward shaping.

\subsubsection*{Scalability and resource demand of foundation models in RL training}
Recent research draw attention to the difficulties of incorporating foundation models straight into the RL training loop, where RL training have (i) \textbf{memory limitations}~\citep{wolczyk2023role,wolczyk2024fine,van2024continual} and foundation models have (ii) \textbf{high inference costs}~\citep{hu2023generalpurposerobotsfoundationmodels,xiao2023robotlearningerafoundation,cao2024surveylargelanguagemodelenhanced} that reduce efficiency on complex tasks.

\textit{Recommendation:} Give priority to modular architectures, such as~\citet{ma2023eureka}. Deploy distilled RL policies for low-latency real-world control and use foundation models mainly for offline learning and reasoning in the RL training loop. Create hybrid pipelines in which smaller controllers make decisions in real time, while LLMs or VLMs offer symbolic reasoning.

\subsection{Challenges in \textit{RL for Generative Policies}}
\label{sec:c2}
\textit{RL pre-training} of generative policies is widely studied and shares challenges with classic RL training as described in prior work~\citep{mnih2013playingatarideepreinforcement,712192,mnih2015human}, we first focus on \textit{RL fine-tuning}, which presents new and specific open challenges unique to generative policies. While, unpredictability, safety verification, and robustness under uncertainty are inherent challenges that are common to both pre-training and fine-tuning. We highlight key open problems in these directions.

\subsubsection*{Policy agnostic RL fine-tuning}
Fine-tuning of generative policies remains a critical challenge for deploying policies in real-world or dynamic environments. The main challenges are: (i) \textbf{obtaining new fine-tuning RL methods} and \textbf{fine-tuning methods independent from the backbone model}~\citep{wolczyk2024fine,ziegler2019fine,chen2025fdpp,dppo2024}, and (ii) \textbf{achieving good fine-tuning for any size of the policy}~\citep{FLaReRL2024}.

\textit{Recommendation:} Investigate new specific methods for efficient RL fine-tuning of VLA models~\citep{FLaReRL2024,mark2024policy}, or smaller diffusion models~\citep{dppo2024}. Develop methods that are agnostic from the policy architecture, such as in~\citet{mark2024policy}.

\subsubsection*{Online RL fine-tuning of VLA models}
Four major obstacles stand in the way of fine-tuning VLA models with RL: (i) \textbf{data gathering is expensive in real-world}, and VLA models are usually trained with real-world data through IL, limiting GPU parallelization in simulation~\citep{kim2024openvlaopensourcevisionlanguageactionmodel,open_x_embodiment_rt_x_2023}. It is (ii) \textbf{computationally demanding to conduct online training}~\citep{FLaReRL2024}, and it is (iii) \textbf{difficult to assess VLA model performance}~\citep{li24simpler}. They also suffer from (iv) \textbf{catastrophic forgetting}~\citep{wolczyk2023role,wolczyk2024fine,van2024continual}.

\textit{Recommendation:} Leverage realistic simulations for massive training and few-shot real-world fine-tuning to close the gap between simulation and reality. Integrate offline RL pipelines to minimize the need for online training. Use distillation to condense huge VLA models into lightweight, task-specific policies that allow for effective RL deployment~\citep{julg2025refined}. Test in simulations the VLA models using frameworks like~\citep{li24simpler}. Combine replay‑based continual RL with regularization to mitigate forgetting~\citep{van2024continual,chan2024offlinetoonline}.

\subsubsection*{Safety and failure modes of generative policies}
Generative policies present serious safety and reliability issues in robotics, due to their black-box nature and absence of clear constraints. They may behave in an unpredictable manner. Challenges are: (i) \textbf{closed-loop verification} and (ii) \textbf{real-time monitoring}~\citep{hu2024harmodt,bucker2023latte,li2024online,wolczyk2024fine,reed2022generalist,chi2023diffusion,janner2022diffuser}.

\textit{Recommendation:} Incorporate safety-aware control modules that monitor and oversee the outputs of generative policies, to enforce physical constraints, and direct policy correction in the face of uncertainty~\citep{AgiaSinhaEtAl2024,pmlr-v9-ross10a,gu2022review}. Create closed-loop systems that use environment feedback to continually test and improve policy behavior, and implement hybrid architectures where generative policies are constrained by explicit safety layers or fallbacks~\citep{perez2024puma}. Apply verification techniques such as control barrier functions~\citep{nakamura2025generalizing}.

\subsection{Future Research Directions}
\label{sec:f1}
We highlight three promising research directions based on our findings.

\subsubsection{RL from human feedback}
Techniques such as RL from Human Feedback (RLHF) and related approaches (e.g., Preference Based RL \citep{PBOSurvey2022}) have been used for aligning large language models and other generative systems to produce outputs that are both useful and aligned with human input\citep{ziegler2019fine,luo2024precise}. Future research should explore RLHF to fine-tune generalist generative policies---such as VLA models. However, evaluating robotic actions remains harder than assessing tasks like chatbot responses.

\subsubsection{Actor-critic foundation models}
A new research direction could be to develop specialized foundation models that act as critics in RL or generate in-context rewards zero-shot in real-world settings for policy training.
Traditional RL depends on manually designed rewards, which can be limiting. Instead, specific foundation models can dynamically assess actions using multi-modal inputs, improving learning efficiency and adaptability with respect to generic LLMs~\citep{ma2023eureka}.

\subsubsection{Constraint-aware generative models with optimal control} Based on recent advances in model-based training of IL policies \citep{perez2024puma,xiao2025safediffuser} and extending our discussion on safety in generative policies (Section~\ref{sec:challenges}), we propose that integrating constraint satisfaction methods from optimal control theory, such as control barrier functions, into RL presents a promising direction for safer and more grounded diffusion policies (as suggested in Figure~\ref{fig:safety}). Moreover, adaptive learning-based techniques that dynamically respond to changing environments can significantly enhance the deployment of diffusion policies across various control tasks, much like model predictive safety filters \citep{wabersich2021predictive}. Our argument is further supported by recent works leveraging diffusion models within a receding horizon framework for control action execution~\citep{wabersich2021predictive,zhao2024vlmpc,hansen2024tdmpc2,xue2024fullordersamplingbasedmpctorquelevel,zhou2024diffusion,qi2025gpc,chi2023diffusion}, which integrate model predictive control frameworks obtaining stronger stability.

\section{Conclusion}
\label{sec:conclusions}
The integration of generative AI models and RL marks a transformative advancement in robotics, enhancing reasoning and adaptability in complex tasks~\citep{chu2023acceleratingRL,wang2024,ma2023eureka,ma2024dreureka,xietext2reward,liu2024rl,Huang2023DiffusionReward,FLaReRL2024}. This synergy combines the extensive knowledge and in-context learning of foundation models, the generative capabilities on multi-modal input fusion of generative AI and the high learning capacity in decision-making of RL in robots. Our review critically examined this integration, with a focus on duality aspects between generative AI and RL for robot action generation; extracting policies from text, video and states information. We first studied how \textit{Generative Tools for RL} consist in a solid portion of research papers where LLMs and VLMs play a major role, and how the union of diffusion models and RL is a very exciting trend in this direction, that deserves further investigation. Moreover, the exploration of how RL can harness foundation world models to enhance model-based RL training further highlights the potential of this integration. Finally, we categorized preliminary work on the other side of our duality investigation---\textit{RL for Generative Policies}---and we focused on combining generative AI with RL in training, fine-tuning, and distilling generative policies. While the opportunities for such a research field are vast, we highlighted remaining challenges, notably in grounding abstract representations of foundation models to physical world concepts---an area needing significant exploration. We concluded that RL-based fine-tuning of robotic generative policies, along with the exploration of failure modes of black-box policies, remains underexplored. While, the scalability and computational demands of these systems pose substantial challenges, especially in resource-constrained environments, limiting the possibility of online training. Addressing these challenges is critical for building fully autonomous robots, controlled by generative AI models able to create policies from multi-modal inputs. Based on our findings, we shared promising research directions for the coming years.

\section*{CRediT authorship contribution statement}
\textbf{Angelo Moroncelli:} Writing – original draft, Writing – review and editing, Visualization, Methodology, Investigation, Software, Formal analysis, Conceptualization. \textbf{Vishal Soni:} Writing – original draft, Writing – review and editing, Visualization, Investigation, Software. \textbf{Marco Forgione:} Writing – review and editing. \textbf{Dario Piga:} Writing – review and editing. \textbf{Blerina Spahiu:} Writing – review and editing, Methodology, Conceptualization, Supervision. \textbf{Loris Roveda:} Writing – review and editing, Supervision, Project administration, Funding acquisition, Conceptualization.

\section*{Declaration of competing interest}
The authors declare that they have no known competing financial interests or personal relationships that could have appeared to influence the work reported in this paper.

\section*{Acknowledgements}
The authors would like to thank the Hasler Foundation for providing funding to carry out our research. Our work was partially supported by the GeneRAI (Generative Robotics AI) project, funded by Hasler Stiftung.

\section*{Data availability}
No data was used for the research described in the article.

\section*{Declaration of generative AI and AI-assisted technologies in the writing process}
During the preparation of this work, the author(s) used ChatGPT in order to check and improve the grammar and clarity of the text. After using this tool/service, the author(s) reviewed and edited the content as needed and take(s) full responsibility for the content of the published article.

\bibliographystyle{elsarticle-harv} 
\bibliography{bibliography}

\end{document}